\documentclass[journal]{IEEEtran}
\ifCLASSINFOpdf
\else
\fi
\usepackage{cite}
\usepackage{graphicx}
\usepackage[cmex10]{amsmath}
\usepackage{mathrsfs}
\usepackage{slashbox}
\usepackage{bm}
\usepackage{textcomp}
\usepackage{subfigure}
\usepackage{amssymb}
\usepackage{booktabs} 

\begin{document}
%
\title{Bag-of-words Representation for Biomedical Time Series Classification}
%
%
%

\author{Jin~Wang,
        Ping~Liu,
        Mary~F.H.She,
        Saeid~Nahavandi,~\IEEEmembership{Senior Member,~IEEE,}
        and Abbas~Kouzani,~\IEEEmembership{Member,~IEEE,}
\thanks{Jin Wang and Mary F.H.She are with the Center for Intelligent Systems Research and the Institute for Frontier Material, Deakin University, Australia. e-mail: jay.wangjin@gmail.}
\thanks{Ping Liu is with Department of Computer Science and Engineering, University of South Carolina, USA.}
\thanks{Saeid Nahavandi is with the Center for Intelligent Systems Research, Deakin University, Australia.}
\thanks{Abbas~Kouzani is with the School of Engineering, Deakin University, Australia.}
\thanks{Manuscript received **; revised **.}}

%
%

\markboth{IEEE TRANSACTIONS ON BIOMEDICAL ENGINEERING,~Vol.~**, No.~**, **}%
{Shell \MakeLowercase{\textit{et al.}}: Bare Demo of IEEEtran.cls for Journals}
%



\maketitle

\begin{abstract}
Automatic analysis of biomedical time series such as electroencephalogram (EEG) and electrocardiographic (ECG) signals has attracted great interest in the community of biomedical engineering due to its important applications in medicine. In this work, a simple yet effective bag-of-words representation that is able to capture both local and global structure similarity information is proposed for biomedical time series representation. In particular, similar to the bag-of-words model used in text document domain, the proposed method treats a time series as a text document and extracts local segments from the time series as words. The biomedical time series is then represented as a histogram of codewords, each entry of which is the count of a codeword appeared in the time series. Although the temporal order of the local segments is ignored, the bag-of-words representation is able to capture high-level structural information because both local and global structural information are well utilized. The performance of the bag-of-words model is validated on three datasets extracted from real EEG and ECG signals. The experimental results demonstrate that the proposed method is not only insensitive to parameters of the bag-of-words model such as local segment length and codebook size, but also robust to noise.

\end{abstract}

\begin{IEEEkeywords}
bag of words, codebook construction, clustering, time series classification.
\end{IEEEkeywords}

%
\IEEEpeerreviewmaketitle

\section{Introduction}
%
%
%
%
\IEEEPARstart{W}{ith} the development of modern technology and reduction of hardware cost, a large amount of biomedical signals such as electroencephalogram (EEG) and electrocardiographic (ECG) are collected every day. These biomedical signals are very useful for monitoring human's physical condition. It is however a challenging task to efficiently and effectively analyze these signals. Traditionally, these signals are manually analyzed by professional experts. However, there are several disadvantages of the manual analysis. Firstly, comparing to the large amount of biomedical signals, the number of professional experts, especially the ones with extensive experience is very limited. Secondly, inspection and monitoring of long-term biomedical signals such as EEG and ECG signals are always very time consuming. It is difficult to keep a high level of concentration during a lengthy inspection, giving rise to an increase in the false hit rate by the operator. Finally, it is frequently needed to find inter-reader variability in the manual inspection and monitoring by experts. Therefore, an automated system that can assist professional experts to analyze long-term biomedical signals is very valuable in real-word applications.

\begin{figure*}[tb]
\centering
\includegraphics[width=5.8in]{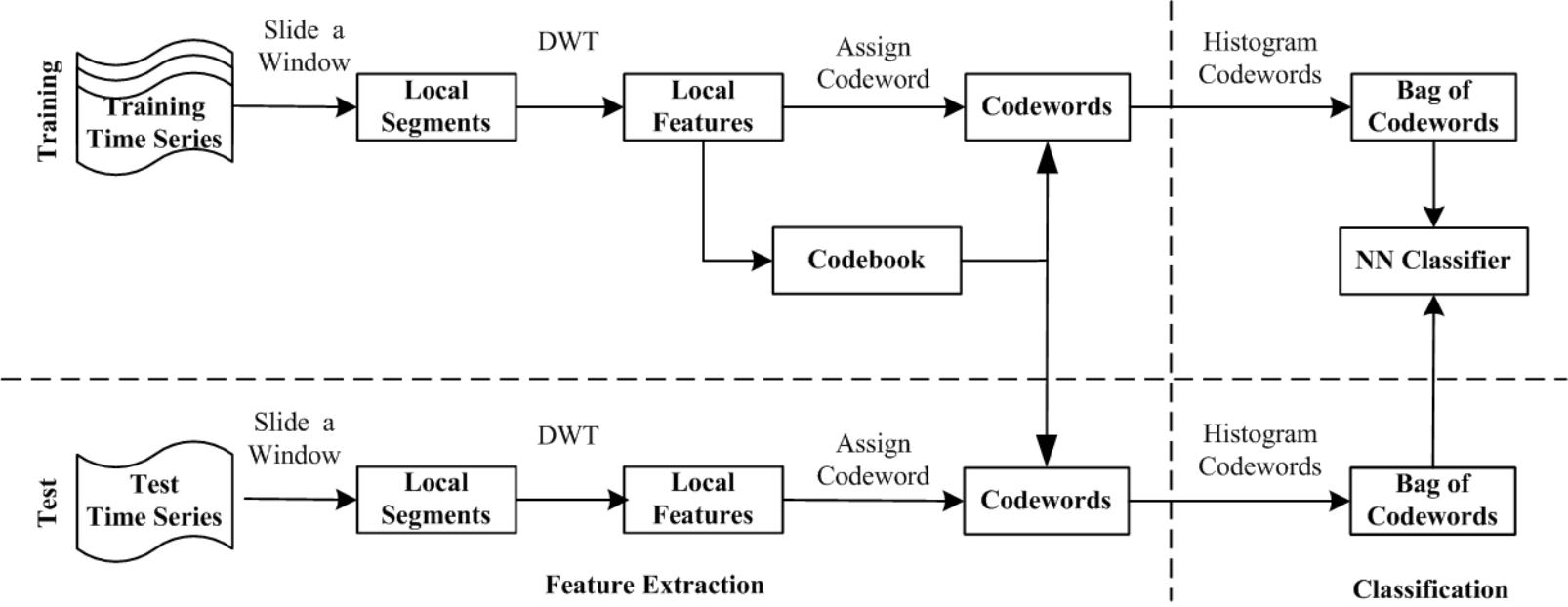}
\caption{The flowchart of the proposed bag-of-words approach for analysis of biomedical time series.}
\label{flow_chart}
\end{figure*}

Automatic analysis of biomedical time series such as EEG and ECG signals based on machine learning techniques has been applied to a variety of real-word applications. For instance, EEG signals are automatically analyzed for epileptic seizure detection \cite{Kannathal2005}\cite{Guo2010}\cite{Yuan2011}, brain computer interaction \cite{Wolpaw2002}\cite{Lu2010}\cite{Wang2011}, human mental fatigue detection \cite{Shen2007} and emotion recognition \cite{Lin2010}. ECG signals that provide useful information about heart rhythm are used to study heart arrhythmias \cite{Ince2009}\cite{Kampouraki2009}. It is essential to extract meaningful features to represent individual time series in the aforementioned applications. Some methods \cite{Huken2003}\cite{Zadeh2010} directly describe time series in time domain while some others extract features from transformed domain \cite{Guo2010}\cite{.I.Guler2005}\cite{Kampouraki2009}. For instance, Zadeh et al. \cite{Zadeh2010} extracted morphological and timing-interval features from ECG segments to classify heartbeats. Guo et al. \cite{Guo2010} extracted line length features based on Discrete Wavelet Transform (DWT) to detect epileptic EEG segments.

Most of the previous representations extract local temporal or frequency information to characterize time series, which are very effective for short time series or time series with periodic waveforms. However, they may have limited ability to capture structural similarity of long time series which have repetitive but un-periodic waveforms, for instance, Electrocardiography (ECG) and Electroencephalography (EEG) signals. In order to capture the high-level structural information of time series, Lin \cite{Lin2012} proposed a bag-of-patterns (BoP) representation by converting a time series to a words string using the Symbolic Aggregate approXimation (SAX). The temporal order of local segments, i.e., local patterns, in a time series is ignored and all the local segments in the time series are histogrammed to construct a bag-of-patterns representation. The bag-of-patterns representation is effective to capture the structural similarity of time series. However, one drawback of the bag-of-patterns representation is that its dimension may be very high, which limits its application for large datasets. For instance, when the size of the alphabet $\tau$ and the number of symbols $w$ are 4 and 8, respectively, the dimension of the bag-of-patterns representation could reach $\tau^{w} =  65536$.

In this work, motivated by the success of the bag-of-words model in text document analysis \cite{Lebanon2007}\cite{Blei2003} and image analysis \cite{Fei-Fei2005}\cite{Niebles2008}, we propose a simple yet effective bag-of-words representation whose dimension is much lower than the bag-of-patterns representation to characterize biomedical time series. The bag-of-words representation is able to capture high-level structural information of time series due to the utilization of both local and global information. Moreover, it can be used to represent streaming data and time series with different lengths because it is incrementally constructed.

The bag-of-words model was originally developed for document representation. The basic idea is to define a codebook that contains a set of codewords and then represent a document as a histogram of the codewords, each entry of which is the count of a codeword occurred in the document. Although the order information of words is ignored, the bag-of-words model is still very effective to capture document information because the frequency information of codewords in documents are well explored. Recently, the bag-of-words model is extended to analyze images and videos in computer vision \cite{Fei-Fei2005}\cite{Niebles2008}. Local patches extracted from images or videos are treated as words and the codebook is constructed by clustering all the local patches in the training data. Similar to the extension of the bag-of-words representation in computer vision, we here extend the bag-of-words representation to characterize biomedical time series by regarding local segments extracted from time series as words and treat the time series as documents.

\subsection{Overview of the Proposed Approach}
In the bag-of-words representation, a time series is treated as a text document and local segments extracted from the time series as words. The general flowchart of the proposed method is demonstrated in Fig. \ref{flow_chart}. Firstly, we continuously slide a window with a pre-defined length along the time series to extract a group of local segments. Then, we extract a feature vector from each of the local segments using DWT. Next, similar to the bag-of-visual-words model in images and videos analysis \cite{Fei-Fei2005}\cite{Niebles2008}, all local segments from the training time series are clustered by k-means clustering to create a codebook, i.e, the cluster centers are treated as codewords. Then, a local segment is assigned the codeword that has the minimum distance to the local segment, and the time series is represented as a histogram of the codewords. Finally, the bag-of-words representation is used as input for classification.

\subsection{Contribution and Organization}
The main contribution of the paper is twofold: (i) a simple yet effective bag-of-words representation is proposed for analysis of biomedical time series; (ii) a series of experiments was conducted to investigate the effectiveness and robustness of the bag-of-words representation for classification of biomedical time series.

The structure of the paper is organized as follows. In Section \ref{sec:method}, the proposed method including the bag-of-words representation, distance measures and classification method is described. Section \ref{sec:dataset} describes the biomedical time series datasets used in the experiments. Experimental results are reported and analyzed in Section \ref{sec:experiment}. Discussion and conclusion are given in Section \ref{sec:discussion} and Section \ref{sec:conclusion}, respectively.

\section{Proposed Method}
\label{sec:method}
In this section, we describe the bag-of-words representation for biomedical time series classification. The bag-of-words representation ignores the temporal order of local segments within a time series and represents the time series as a histogram of codewords i.e., local segments. Several distance measures are then introduced for the histograms comparison.

\subsection{Bag-of-words Representation}
\begin{figure*}[tb]
\centering
\includegraphics[width=6in]{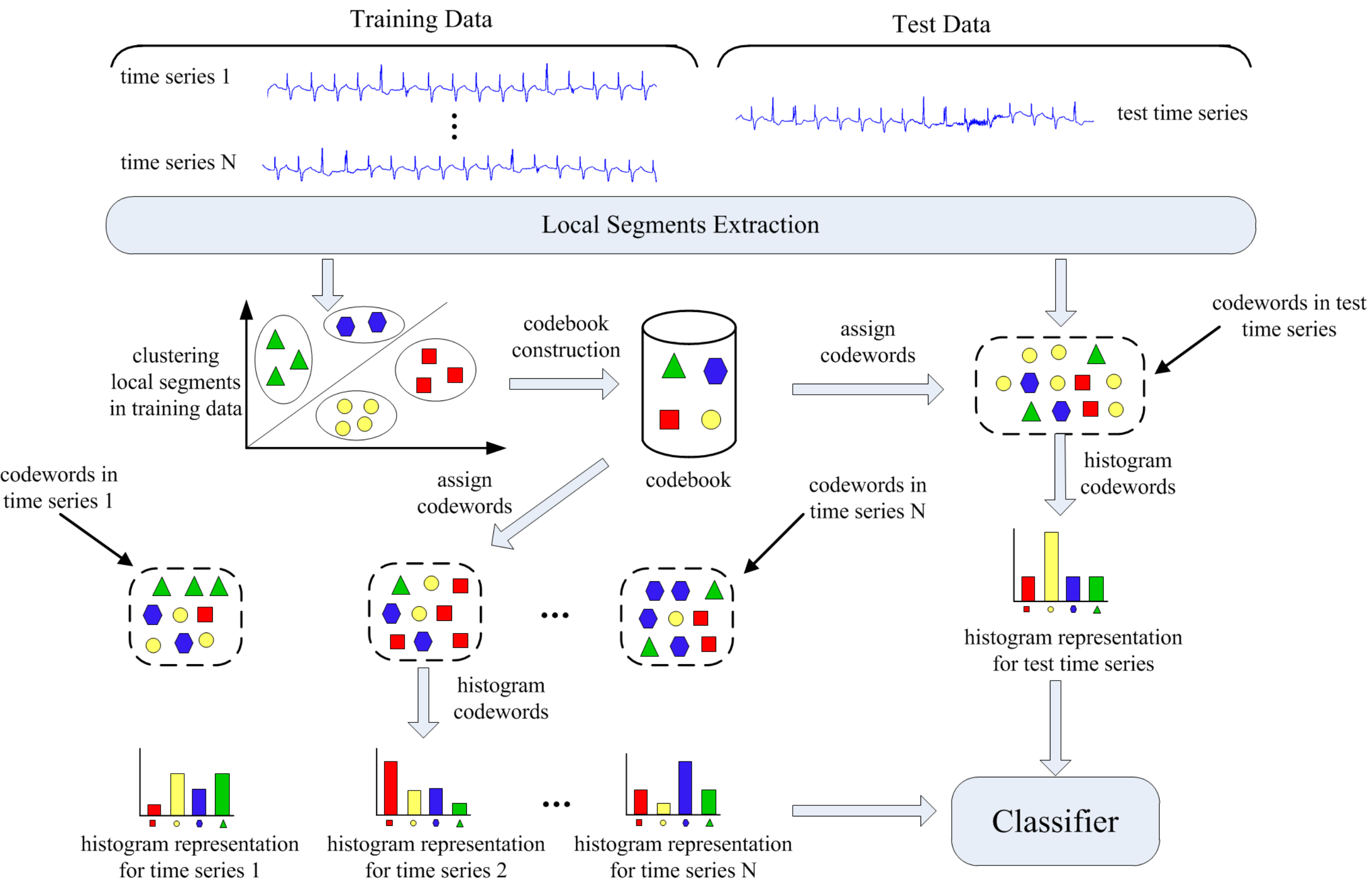}
\caption{The procedure of generating the bag-of-words representation. The codebook is constructed by clustering all local segments from training data. The ``circle'', ``triangular'' ``square'' and ``hexagon'' stand for the basis elements, i.e., codewords, in the codebook. Each local segment is assigned a codeword and a histogram representation is extracted for each time series by histogram the codewords in the time series. The figure is best viewed in color.}
\label{bag_of_words}
\end{figure*}

The procedure of generating the bag-of-words representation is illustrated in Fig. \ref{bag_of_words}. We continuously slide a window with pre-defined length along a time series and extract a group of local segments from the time series. A feature vector is then extracted from each of the local segments using the DWT to characterize the local segment. All the local segments from the training data are clustered to construct a codebook that contains a set of codewords, i.e., the cluster centers. Then, a local segment is assigned the codeword that has minimum distance with the local segment. The bag-of-words representation ignores the order of local segments in a time series and represent the time series as a histogram of codewords.

\begin{figure*}[tb]
\centering
\includegraphics[width=6in]{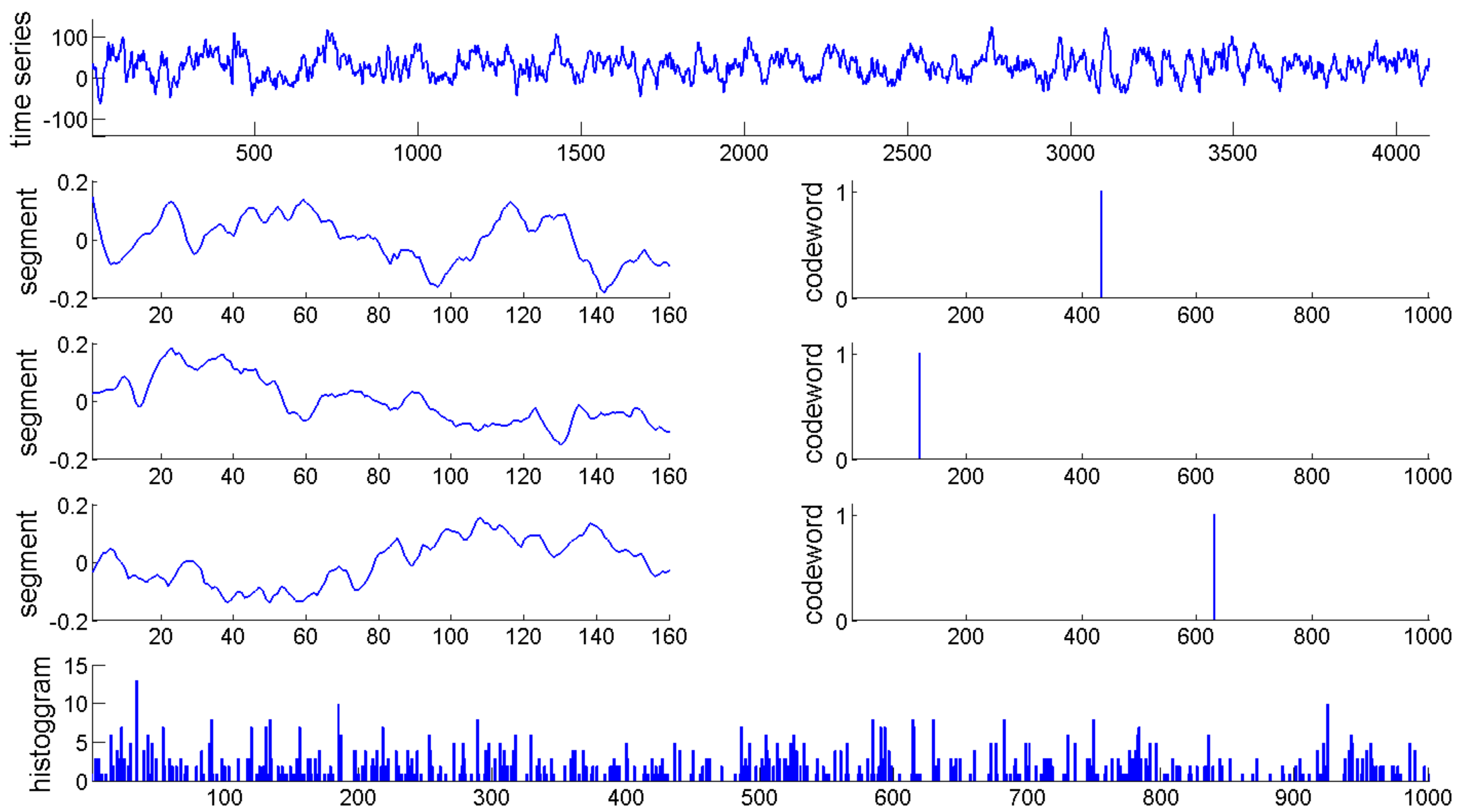}
\caption{The bag-of-words representation of an example time series. See the corresponding text for more details.}
\label{Fig_BoW}
\end{figure*}

\subsubsection{Local Segments Extraction}
A group of local segments are extracted from each time series by continuously sliding a window with pre-defined length along the time series. As local segments from different time series may be at different scales, all the local segments are normalized to zero mean and standard deviation. We transform a local segments into wavelet domain and use  approximations wavelet coefficients of DWT as a feature vector to represent the local segment.

The wavelet transform that analyzes a signal at different frequency bands provides both accurate frequency information at low frequencies and time information at high frequencies, which are very important for biomedical signal analysis. The choice of wavelet function and the number of decomposition levels is of importance for the multiresolution decomposition. In this work, a single level DWT with order 3 Daubechies wavelet function (db3) is employed to decompose a local segment into approximations coefficients and detailed coefficients. Similar to the work in \cite{.I.Guler2005}, we used the approximation coefficients as a feature vector to represent the local segment. We do not directly use the raw value of local segments as feature vectors due to the fact that features using the approximations coefficients not only are more robust to noise than features using raw segments but also have nearly half dimension of the local segments.

\subsubsection{Codebook Formulation}
In the text document analysis, a codebook (vocabulary) is a set of pre-defined words, which are also called codewords. The bag-of-words method counts the number of each codeword that exists in a document and provides a document-level representation using a histogram of codewords. In image and video analysis, such codebook is generally created by performing clustering on a group of local patches from training data, i.e., the codewords are defined as the clustering centers. The codeword that is nearest to a local patch is then assigned to the local patch. The spatial and temporal order information of local patches (codewords) is ignored and an image or video is represented as a histogram of codewords in the image or video. The classical k-means clustering algorithm \cite{Fei-Fei2005}\cite{Niebles2008} is commonly used to construct the codebook, although some other unsupervised and supervised methods are also developed such as mean-sift \cite{Jurie2005} and supervised Gaussian mixture models \cite{Fernando2012}.


Similar to the codebook construction in image and video analysis, we cluster all the local segments from training time series using k-means clustering to construct the codebook. The clustering centers estimated by the k-means clustering are regarded as basis elements of the codebook, i.e., codewords. Suppose a group of local segments $\mathbf{X}=[\mathbf{x}_1, \mathbf{x}_2, \cdots, \mathbf{x}_n] $, where $\mathbf{x}_i \in \mathbb{R}^{d}$, are extracted from taring time series, the codebook construction by k-means clustering is formulized as the optimization problem:

\begin{equation}
\begin{split}
 \label{equ:kmeans}
 \min_{\mathbf{B}\in \mathbb{R}^{d\times K},\mathbf{V}\in \mathbb{R}^{K\times n}} \sum_{i=1}^n \lVert \mathbf{x}_i - \mathbf{B} \mathbf{v}_i \rVert_2,  \\
 s.t. \quad card(\mathbf v_i) = 1 ,\lvert \mathbf v_i \rvert = 1, \forall i, \mathbf v_i \geq 0,
 \end{split}
\end{equation}
where $\mathbf{B}\in \mathbb{R}^{d\times K}$ is the clustering centers and the vector $\mathbf{v_i}$ is the clustering index of the local segment $\mathbf{x_i}$, which is a unit-basis vector that has only one component equal to one and all the other components are zero. The codebook $\mathbf{B}\in \mathbb{R}^{d\times K}$ has $K$ codewords, each of which is a $d$-length vector, the same length as the local segments. It is worth noting that the codebook only needs to be learned once from training data and it is universal for both training and test data.

The codebook size $K$ is of importance to the bag-of-words representation. A compact codebook with too few entries has a limited discriminative ability, while a large codebook is likely to introduce noise due to the sparsity of the codewords histogram. Therefore, the size of the codebook should well balance the trade-off between discrimination and noise.

\subsubsection{Codewords Assignment}
Once the codebook is constructed, a local segment is assigned the codeword that has minimum distance with the local segment. Specifically, suppose that a codebook with $K$ entries, $\mathbf{B}=\{\mathbf{b}_1,\mathbf{b}_2,...,\mathbf{b}_K\}$, is learned from the training data. A local segment $\mathbf{x}_i$ is assigned the $c$ th codeword that: \mbox{$c^{*}=\arg\min_j d(\mathbf{b}_j, \mathbf{x}_i)$}, where $d(\cdot, \cdot)$ is the Euclidean distance function.

After each local segment is assigned a codeword, the temporal order of local segments is ignored and a time series is represented as a histogram of codewords in the time series, each entry of which specifies the count of a codeword occurred in the time series. Fig. \ref{Fig_BoW} illustrates the bag-of-words representation of an example EEG time series. The figure in the first row is the example EEG time series. The three figures in the second to fourth rows (left) are three local segments with length of 160 extracted from the time series, and the three figures in the second to fourth rows (right) are the corresponding codewords assigned to the local segments from codebook, which consists of 1000 codewords. The three local segments are assigned the $432 th$, $118 th$, and $628 th$ codewords, respectively. The figure in the last row is the bag-of-words representation for the time series, each entry of which gives the count of a codeword occurred in the time series.

%
%
%
%
%

\subsection{Classifier}
Some discriminative classifiers such as Artificial Neural Networks (ANN) \cite{Guo2010}, Support Vector Machine (SVM) \cite{I.Guler2007}, and Probabilistic Neural Networks (PNN) \cite{.I.Guler2005} are widely used for biomedical signal classification. Since our goal in this paper is to investigate the effectiveness of the bag-of-words representation, here we use the simplest classifier, i.e., the \mbox{1-Nearest Neighbor} (1-NN) classifier. Let $\mathbf{t}$ be a test time series and $\mathbf{R}^i$ represents the time series from the $i$th category. The test data is determined as the class $C$ of the training sample that has minimal distance with the test data, i.e., \mbox{$C^{*}=\arg\min_i D(\mathbf{t},\mathbf{R}^i)$}, where $D(\cdot,\cdot)$ is the similarity measure that is defined in the following.

\subsection{Similarity Measure}
\label{sec:distance}
Many similarity measures have been proposed for histograms comparison. In the following, we describe four commonly used similarity measures for distance measurement of two bag-of-words representations.


\subsubsection{Euclidean Distance}
The Euclidean distance between histogram $\mathbf{h}$ and histogram $\mathbf{k}$ is defined as:
\begin{equation}
 \label{equation 3}
   D_{L_2}(\mathbf{h},\mathbf{k})=\left(\sum_{i}\left | h(i)-k(i) \right|^2\right)^{1/2},
\end{equation}
where $D_{L_2}(\mathbf{h},\mathbf{k})$ is the Euclidean distance, which is commonly used in pattern recognition.

\subsubsection{Chi-Squared Distance}
The Euclidean distance subtracts the two histograms bin-by-bin and contributes each bin pairs equally to the distance. The problem is that some words such as ``the", ``but" and ``however" occur more frequently in documents; therefore, they contribute more to the distance in the Euclidean Distance measure. But they may actually have less discriminative information than rarely happened codewords. This leads to the Chi-Squared distance ($\chi^2$ distance):
\begin{equation}
 \label{equation 4}
   D_{\chi^2}(\mathbf{h},\mathbf{k})=\sum_{i}\frac{\left | h(i)-k(i) \right|^2}{h(i)+k(i)+\varepsilon},
\end{equation}
where $\varepsilon$ is a small value to avoid dividing by zero. The $\chi^2$ distance introduces a normalization to emphasis the rarely happened codewords because common words are always shared between documents from different categories.

\subsubsection{Jensen-Shannon Distance}
Each entry of the bag-of-words represents can be interpreted as the frequency of a codeword occurred in a time series. Therefore, the histogram stands for a probabilistic distribution over discrete random variables. A simple measure to compare two distribution is the Kullback-Leibler divergence:
\begin{equation}
 \label{equation 5}
   D_{KL}(\mathbf{h}||\mathbf{k})=\sum_{i}h(i)\left( log_2h(i)-log_2k(i)\right).
\end{equation}
If and only if $\mathbf{h}$ and $\mathbf{k}$ are the same, the KL divergence becomes zero. In order to keep the distance symmetric,  the Jensen-Shannon distance \cite{Endres2003} is introduced as a symmetric extension:

\begin{equation}
 \label{equation 6}
   D_{JS}=\frac{1}{2}\left(D_{KL}(\mathbf{h}||\mathbf{k})+D_{KL}(\mathbf{k}||\mathbf{h})\right).
\end{equation}

\subsubsection{Histogram Intersection based Distance}
The histogram intersection which counts the total overlap between two histograms is able to address the problem of partial matches when the two histograms have different sum over all the bins. The distance based on the histogram intersection is defined as \cite{Grauman2007}:
\begin{equation}
 \label{equation 8}
   D_{HI}(\mathbf{h}||\mathbf{k})=1-\sum_{i}\max \left( h(i),k(i)\right),
\end{equation}
where $\mathbf{h}$ and $\mathbf{k}$ are normalized histogram vectors. Two histograms that have larger overlap will obtain a smaller distance.

\subsection{Practical Implementation}
A large number of local segments may be extracted from training data, especially for large datasets. Clustering a large number of local segments to construct the codebook is computationally expensive. In practice, instead of using all the local segments extracted from the training data, we performing the k-means clustering on a subset of local segments randomly selected from the training data to construct the codebook. This strategy is also employed in image and video analysis to reduce the computation of codebook construction \cite{Fei-Fei2005,Niebles2008}.

We continuously slide a window along a time series to extract local segments. However, when the time series contains too many data points, a large number of local segments will be extracted from the time series, which requires expensive computation. For instance, for a time series consisting of 2000 data points, about 1900 local segments will be extracted using a window with 100 length. In practice, we can slide the window with a step of $n$ data points ($n=2,4,6$ or $8$) along the time series to reduce the number of local segments extracted from the time series.

The MATLAB code of the bag-of-words representation in this work was made publicly available at http://www.mathworks.com/matlabcentral/fileexchange/38050.

\section{Experimental Datasets}
\label{sec:dataset}
In this study, three datasets constructed from EEG and ECG signals are used to evaluate the performance of the bag-of-words representation. The first dataset is collected from EEG signals and it is widely used for automatic epileptic seizure detection. The other two datasets are extracted from long ECG signals (more than 1000000 points) that collected from different subjects with random start points. Each of the long ECG signals corresponds to a class, i.e., subjects' identity. In order to demonstrate that the bag-of-words representation can be for time series with different lengths, the third dataset is extracted with different lengths between 2048$\sim$4096, while the first two datasets have the same length of 4096 and 2048, respectively.

It is worth noting that although the extracted ECG time series in the same class are obtained from the same long ECG signal, there exist substantial inter-class variations. The aim of the ECG signal classification in our experiment is to attribute each instance, i.e., extracted ECG time series, to their subjects' identity, which can be used for human identification from ECG signals in real application \cite{Biel2001,Fang2009}. This task may be not difficult by comparing features extracted based on heartbeat waveforms or fiducial points, for instance, cross-correlation among QRS complexes. However, the bag-of-words representation does not need to detect or localize any heartbeat waveforms or fiducial points, which is always required in previous works \cite{Biel2001,Fang2009}.

\begin{table} [!t]
 \renewcommand{\arraystretch}{1.3}
  \caption{The three datasets used in the experiments.}
  \label {datasets}
  \centering
  \begin {tabular} {l r r r}
  \toprule[1pt]
   \bfseries Datasets &\bfseries Classes &\bfseries Num of Sequences &\bfseries Length of sequences \\ \midrule
    EEG     &5  &500   &4096 \\
    ECG-40  &40 &2000  &2048 \\
    ECG-15  &15 &1500  &2048$\sim$4096 \\
  \bottomrule
  \end{tabular}
\end{table}

\subsection{EEG Dataset}
\begin{figure}[tb]
\begin{center}
\includegraphics[width=3.4in]{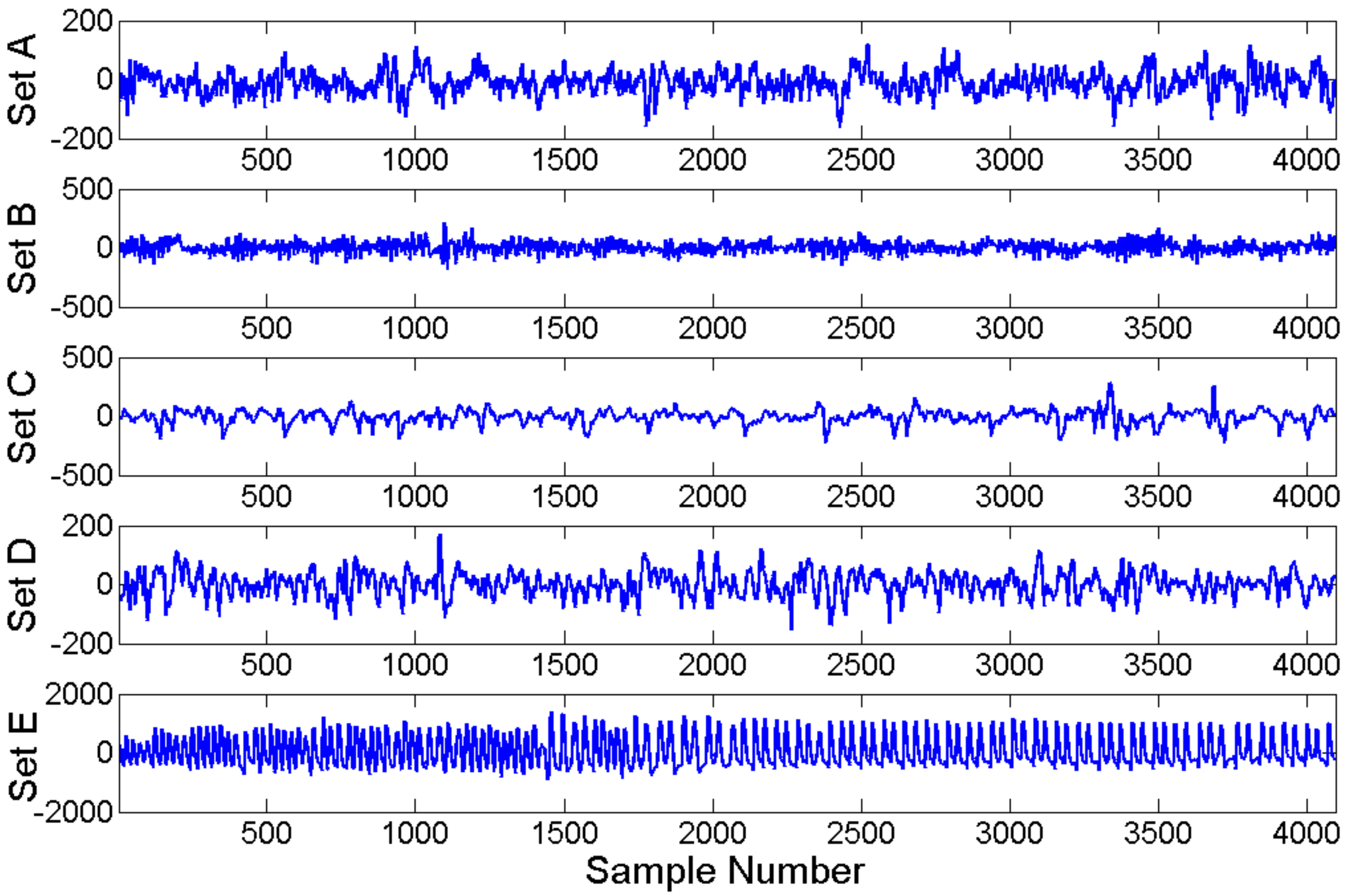}
\caption{Example EEG sequences from each of the five classes.}
\label{datasetEEG}
\end{center}
\end{figure}

The EEG dataset described in \cite {Andrzejak2001} was used in our experiments. The complete EEG dataset consists of five classes (i.e., A, B, C, D, and E), each of which contains 100 single-channel EEG sequences of the same length 4096. All the signals were recorded with the same 128-channel amplifier system and visual inspected for artifacts. Set A and set B are collected from surface EEG recordings of five healthy subjects with eye open and eye closed, respectively. The other three sets (C, D and E) are taken from intracranial EEG recording of five patients suffered from epileptic. Set C and set D are taken from the epileptogenic zone and the hippocampal formation of the opposite hemisphere of the brain, respectively. Set C and set D were recorded in seizure-free intervals, whereas set E only contains seizure activity. Fig. \ref{datasetEEG} shows example time series from each of the five classes.

\subsection{ECG-40 Dataset}
The ECG-40 dataset was obtained from the Fantasia ECG database \cite{Goldberger2000}. The database consists of twenty youth and twenty old healthy subjects. Forty long ECG signals are collected from each of the forty subjects monitored for about two hours with a sampling rate of 250 Hz. All the signals contain more than 1000000 data points, which are very long. We extracted fifty time series of length 2048 from each of the forty long signals with random start points. Totally, the ECG-40 dataset contains 2000 time series of length 2048, which are evenly distributed in the forty classes.

\subsection{ECG-15 Dataset}
The ECG-15 dataset consists of 1500 time series extracted from fifteen long ECG signals in the BIDMC Congestive Heart Failure Database \cite{Goldberger2000}. The fifteen long ECG signals were recorded from fifteen patients suffered from severe congestive heart failure. One hundred time series of length between 2048 and 4096 are extracted from each of the fifteen long ECG signals with random start points. Totally, the ECG-15 dataset consists of fifteen classes, each of which has 100 time series of length between 2048 and 4096.

Table \ref{datasets} summaries the three datasets used in the experiments. It should be noted that the lengths of the 1500 time series in the ECG-15 dataset are not the same, which vary between 2048 and 4096.

\section{Results}
\label{sec:experiment}
In this section, we report experimental results on the three datasets. Firstly, we investigated the impact of parameters by varying the length of local segments and the size of codebook $K$ based on different distance measures. Then, we compared the proposed method with the Discrete Wavelet Transform (DWT) \cite{.I.Guler2005} representation, the Discrete Fourier Transform (DFT) \cite{Rafiei2000} representation, the NN classifier based on Dynamic Time Warping (DTW) \cite{Fu2011} distance and the bag-of-patterns representation (BoP)\cite{Lin2012}. In addition, we compared the classification accuracies achieved by the proposed method with those achieved by other state-of-the-art methods on the EEG dataset. Finally, we investigated the robustness of the bag-of-words representation to noise. In order to ensure an un-biased evaluation, a dataset is randomly partitioned into $10$ subsets. $Nine$ subsets are used for training while the remaining one is retained for test. The classification process is then repeated $10$ times with each of the $10$ subsets used exactly once as test data.

\subsection{Length of Local Segments}

\begin{figure*}[tb]
\centering
\subfigure[]
 {\includegraphics[width=2.2in]{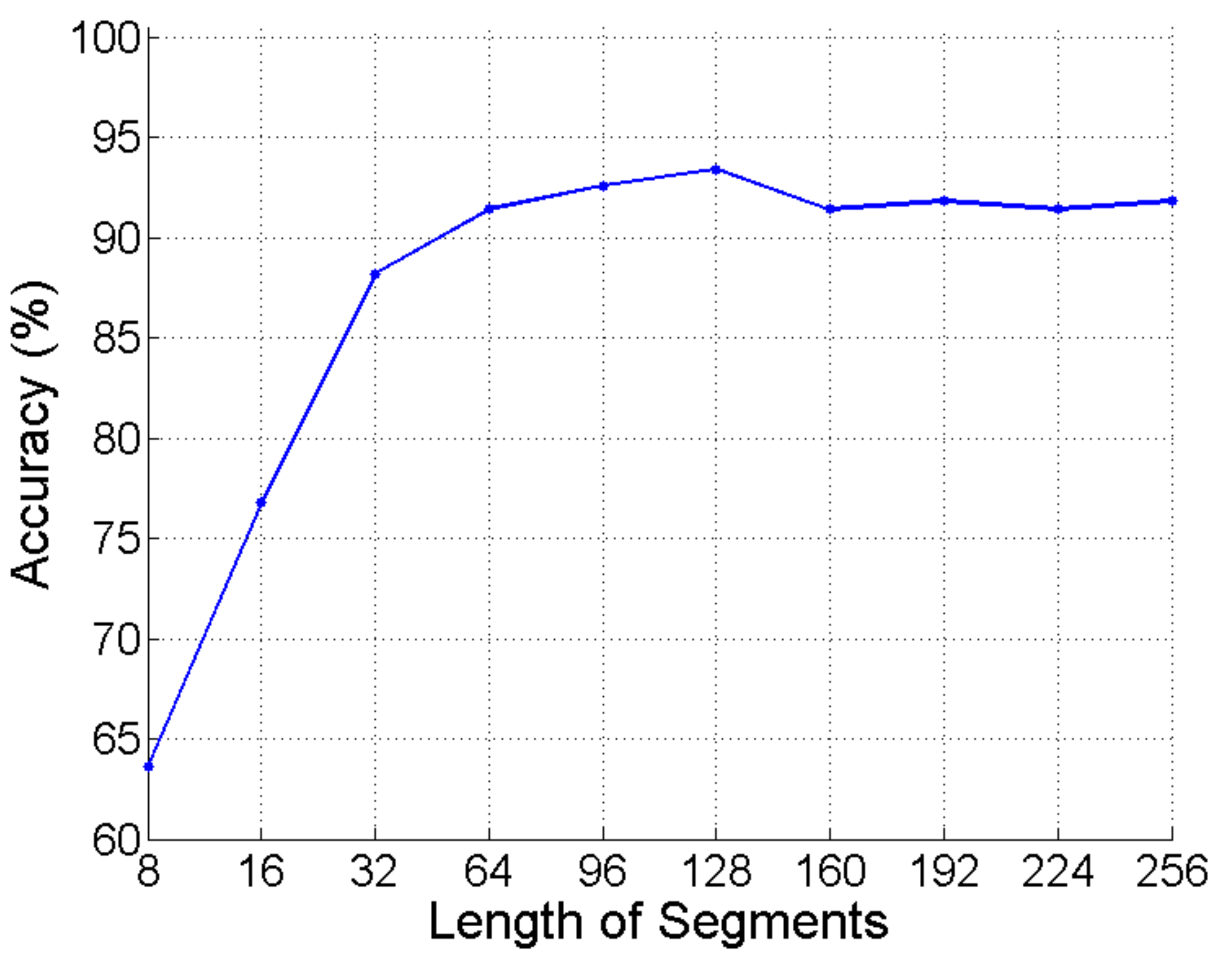}
 \label{length_1}}
\hspace{0.05in}
\subfigure[]
 {\includegraphics[width=2.2in]{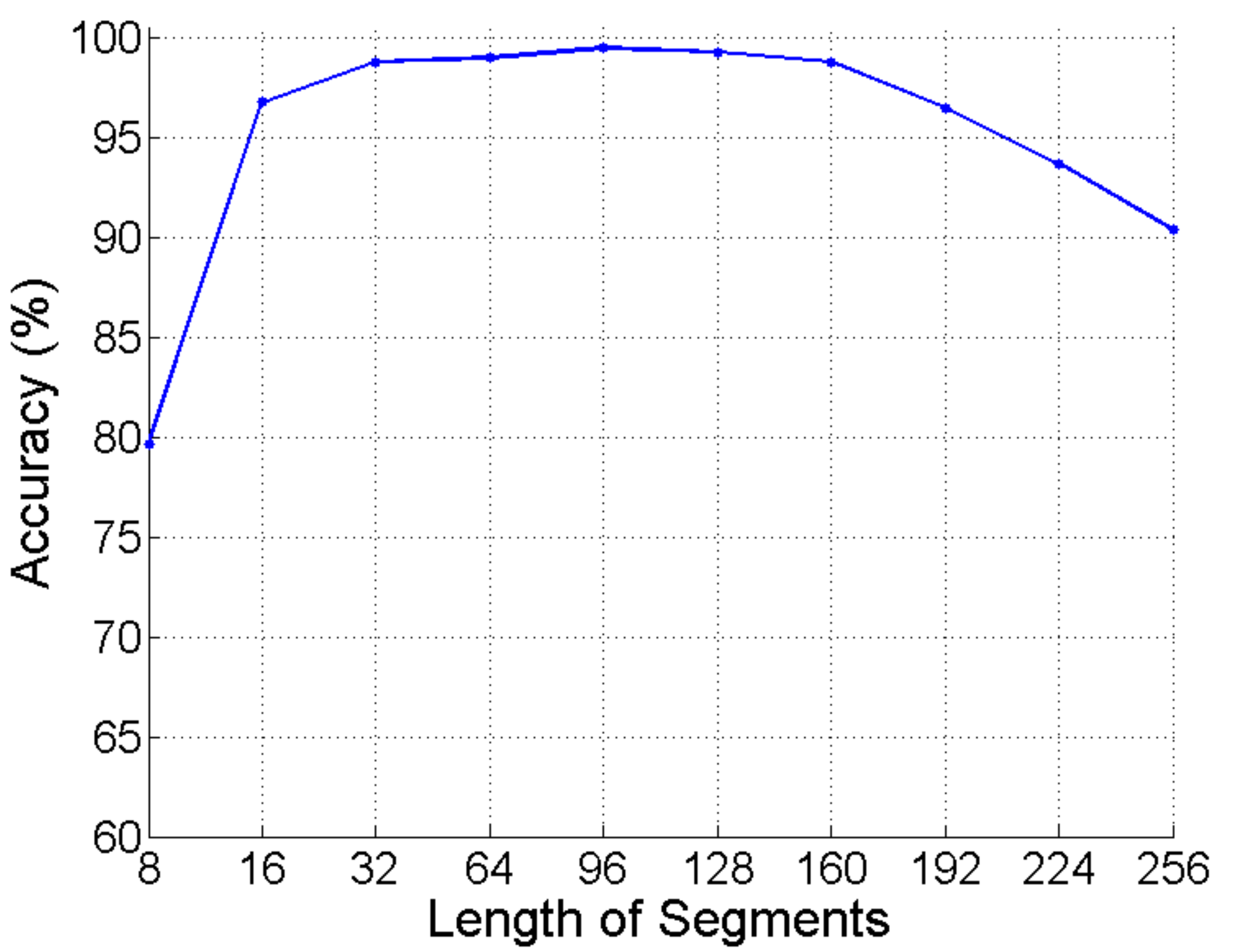}
 \label{length_2}}
 \hspace{0.05in}
\subfigure[]
 {\includegraphics[width=2.2in]{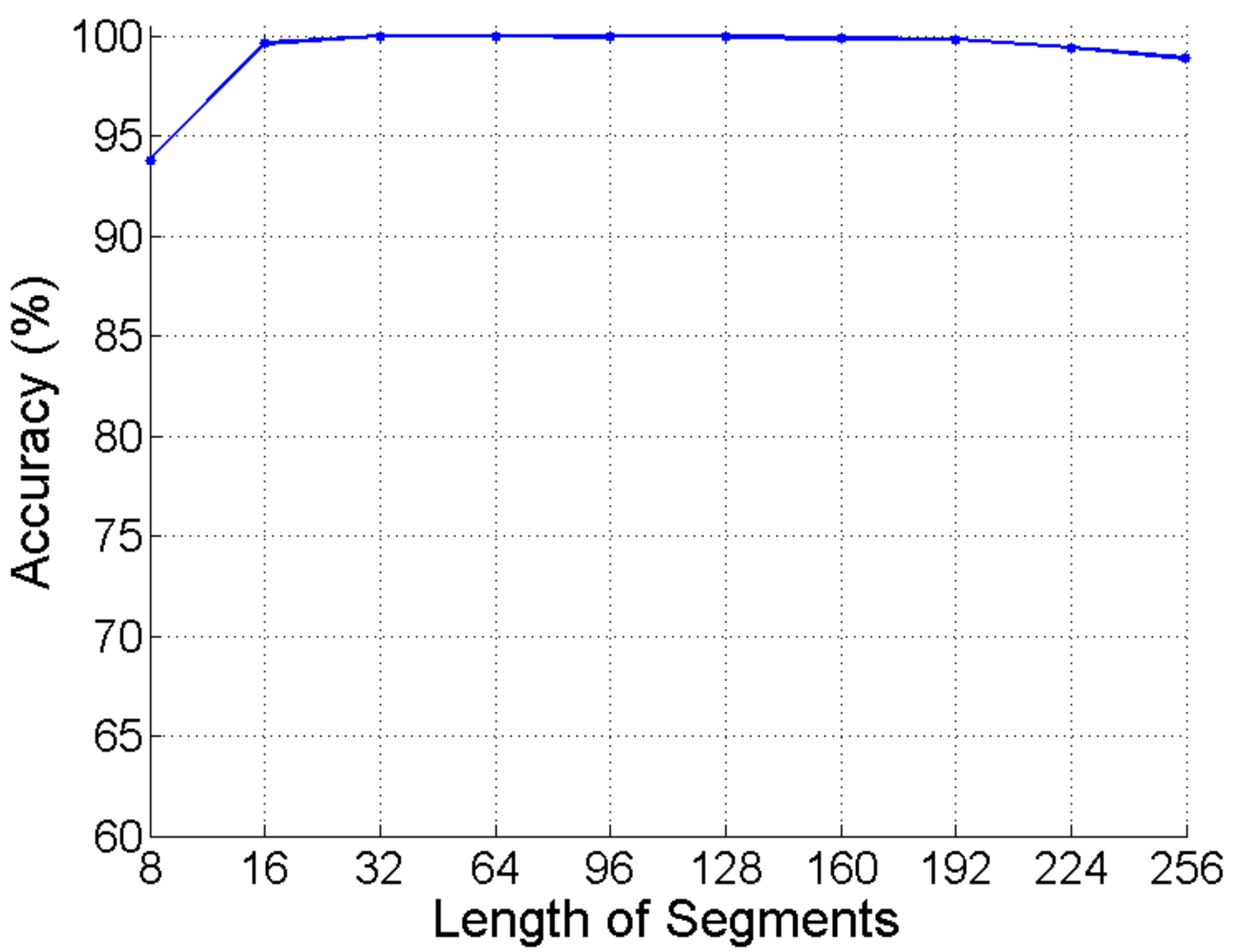}
 \label{length_3}}
\caption{Classification accuracies with respect to the length of segments on the EEG (a), ECG-40 (b) and ECG-15 (c) datasets, respectively.}
\label{length}
\end{figure*}

\begin{figure*}[tb]
\centering
\subfigure[]
 {\includegraphics[width=2.2in]{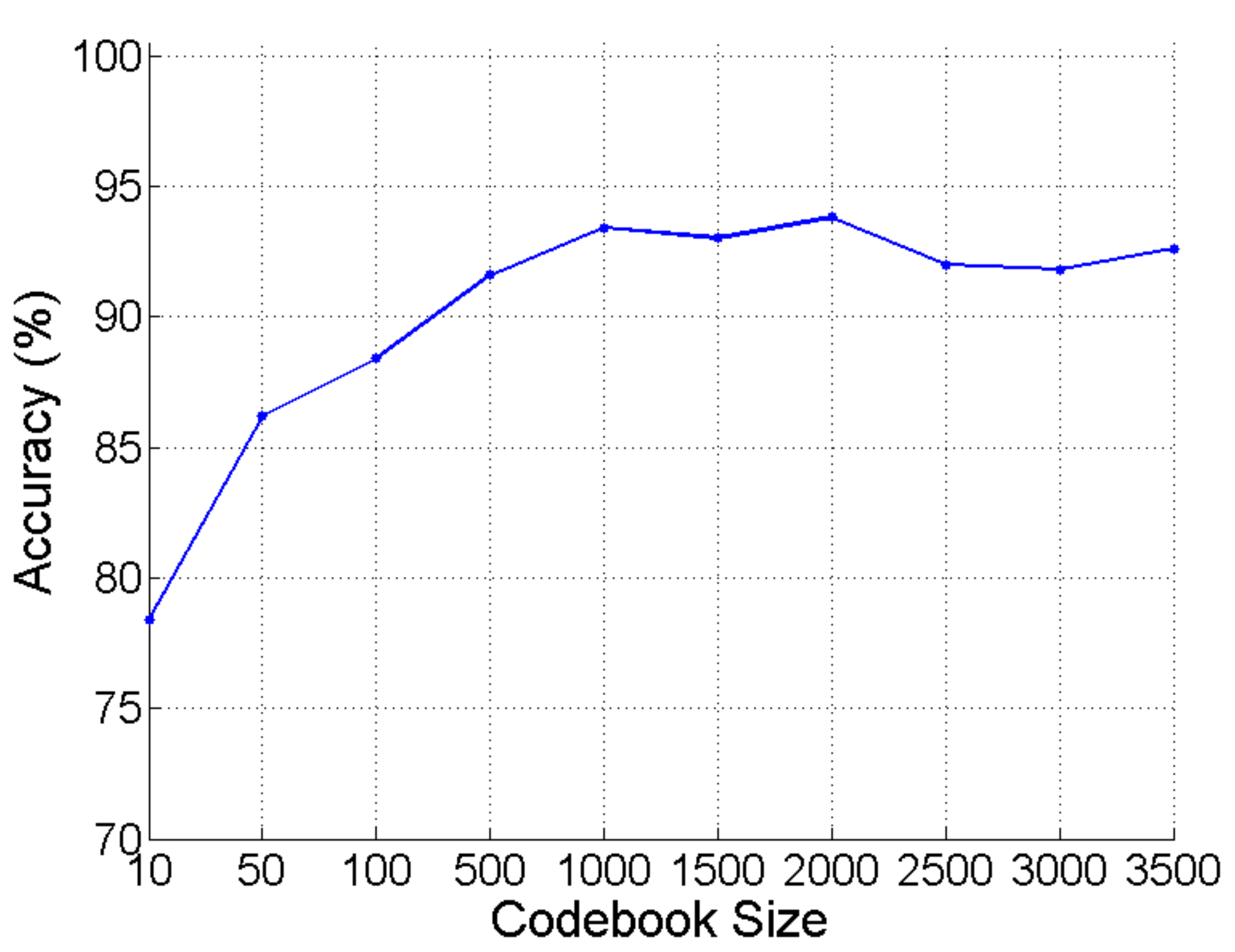}
 \label{codebook_1}}
\hspace{0.05in}
\subfigure[]
 {\includegraphics[width=2.2in]{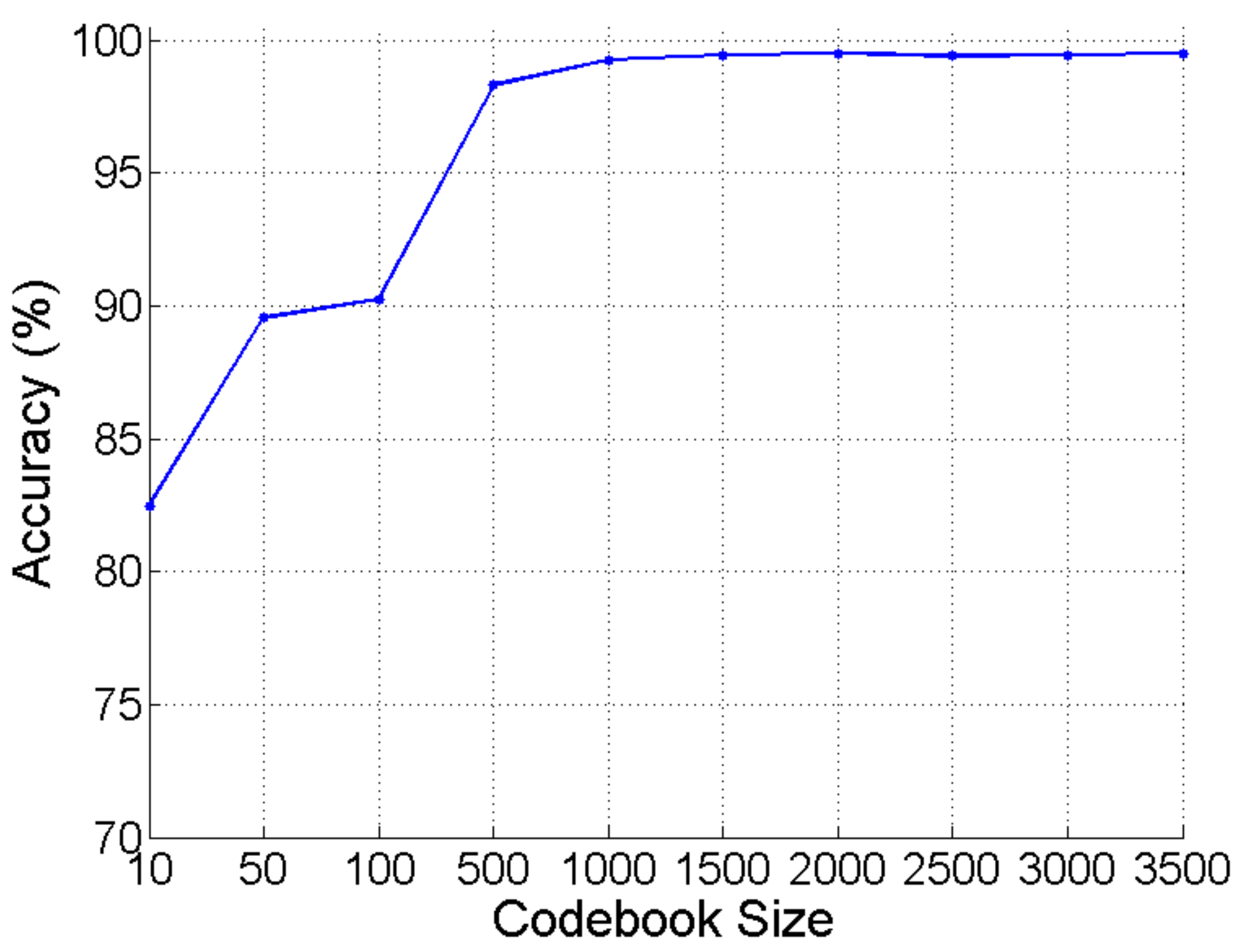}
 \label{codebook_2}}
 \hspace{0.05in}
\subfigure[]
 {\includegraphics[width=2.2in]{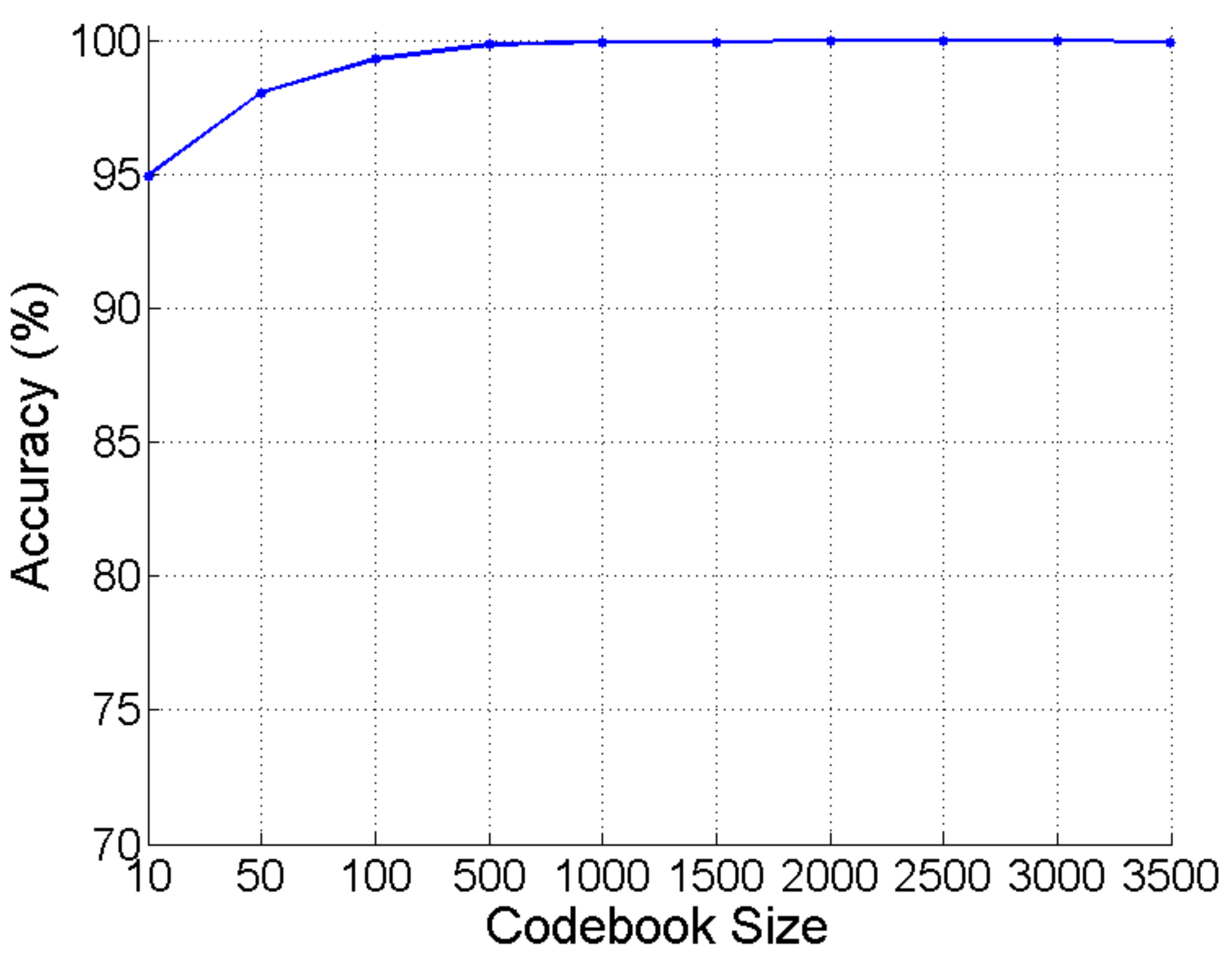}
 \label{codebook_3}}
\caption{Classification accuracies with respect to the codebook size on the EEG (a), ECG-40 (b) and ECG-15 (c) datasets, respectively.}
\label{codebook}
\end{figure*}

\begin{figure*}[tb]
\centering
\subfigure[]
 {\includegraphics[width=2.2in]{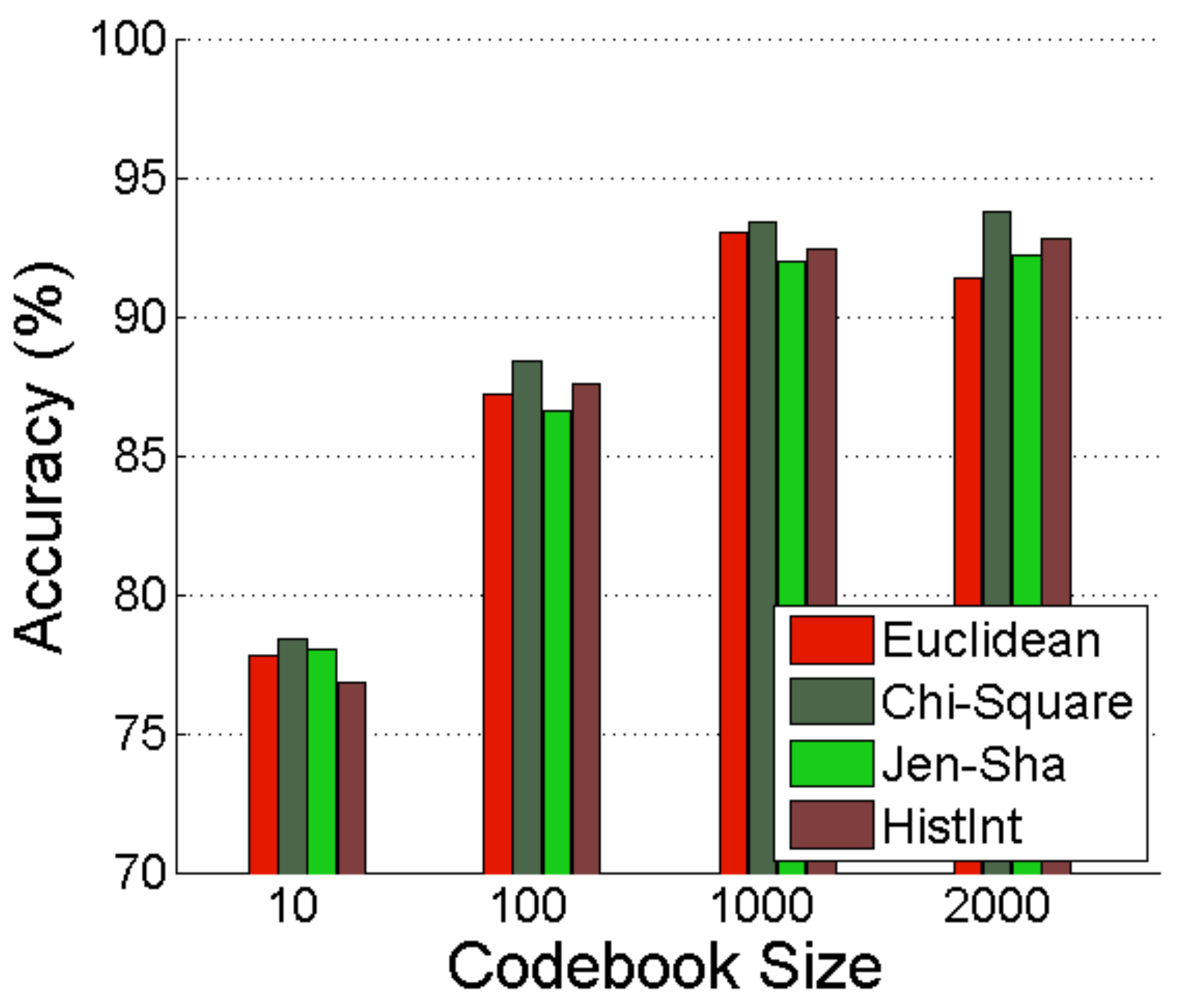}
 \label{distance_1}}
\hspace{0.05in}
\subfigure[]
 {\includegraphics[width=2.2in]{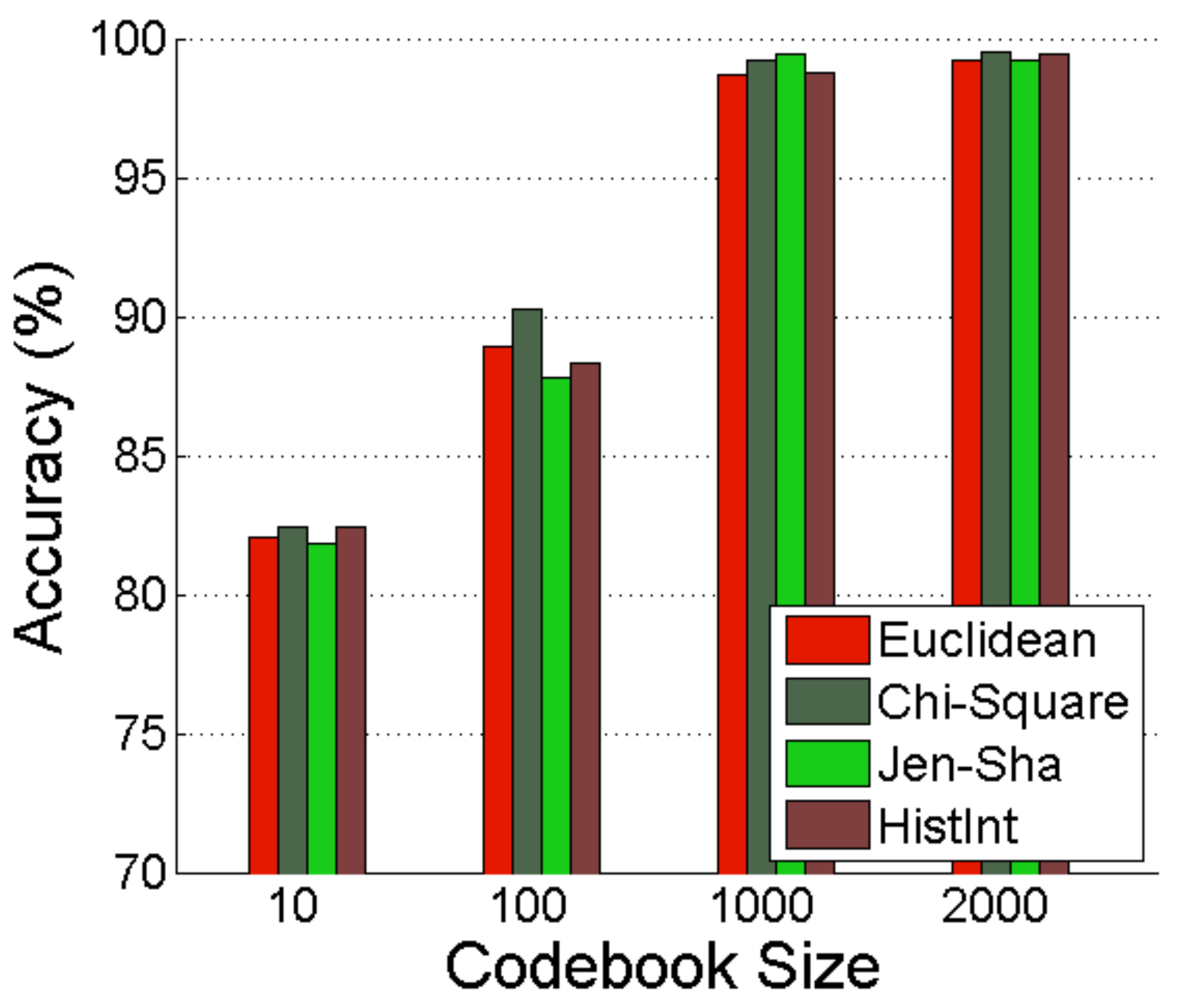}
 \label{distance_2}}
 \hspace{0.05in}
\subfigure[]
 {\includegraphics[width=2.2in]{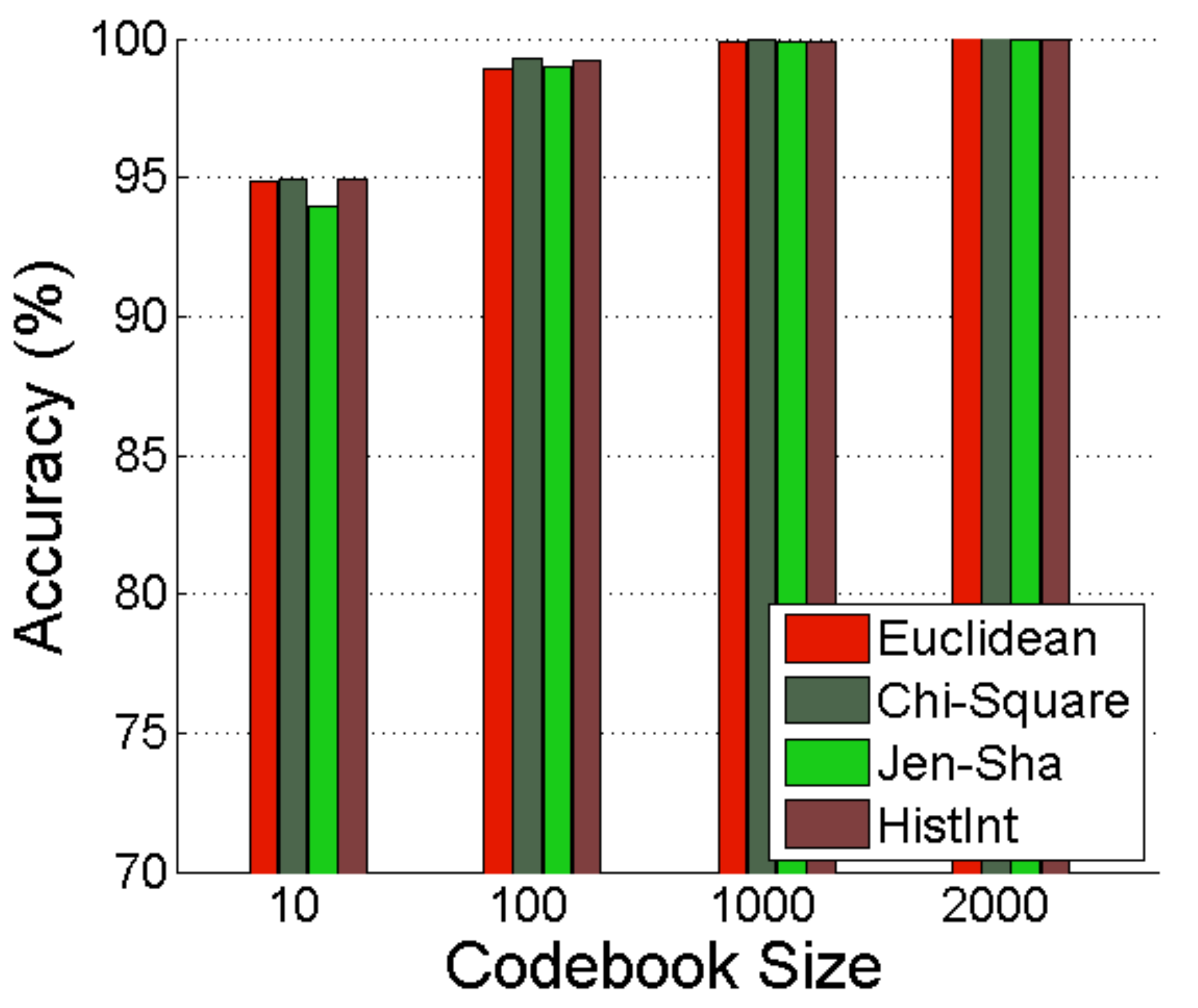}
 \label{distance_3}}
\caption{Classification accuracies using different distance measures on the EEG (a), ECG-40 (b) and ECG-15 (c) datasets, respectively. The figure is best viewed in color.}
\label{distance}
\end{figure*}

We varied the length of local segments between 8 and 256 in the experiments. The determination of such parameter ranges relies on the fact that the biomedical time series such as ECG and EEG signals are relatively flat. The classification accuracies on the EEG, ECG-40 and ECG-15 datasets with a codebook size of 1000 using the Chi-Squared distance is illustrated in Fig. \ref{length_1}, Fig. \ref{length_2} and Fig. \ref{length_3}, respectively. From the experimental results, it can be seen that the performance is relatively stable with respect to the length of local segments when it is between 64 and 192. The classification accuracies decrease considerably with the length less than 16. This is mainly due to the fact that a local segment with too short or too long length can not capture local structure information within time series. In the following experiments, we empirically set the length of local segments as 128.

\subsection{Codebook Size}
To show the performance of the bag-of-words representation with respect to the size of the codebook, we report the classification accuracies on the three datasets in Fig. \ref{codebook}, increasing the size of the codebook from 10 to 3500. We can see that the results become very stable when the size of the codebook is larger than 500. The classification accuracies reduce quickly if the size of the codebook is less than 100, which confirms that a compact codebook with too few entries has a limited discriminative ability. The optimal size of the codebook can be roughly identified as 1000$\sim$3500.

\subsection{Distance Measurement}
We compared the classification performance on the three datasets using the four similarity measures described in Section \ref{sec:distance}. Fig. \ref{distance} demonstrates the classification accuracies based on the four distance measures with the codebook size of 10, 100, 1000 and 2000. We can see that the results are slightly different using various distance measures, indicating that the distance measures have limited impact on the performance of the bag-of-words representation. Overall, the Chi-Squared distance measure performs slightly better than the other measures for all the four sizes of the codebook.

\subsection{Comparison with Other Methods}

\begin{table} [!t]
 \renewcommand{\arraystretch}{1.3}
  \caption{comparison of results on the three datasets using different methods.}
  \label {methods}
  \centering
  \begin {tabular} {l r r r}
  \toprule[1pt]
   \bfseries methods &\bfseries EEG &\bfseries ECG-40 &\bfseries ECG-15 \\ \midrule
    DWT     &76.0  &25.1   &20.1 \\
    DFT     &91.6 &85.6  &60.6 \\
    DTW     &71.6 &74.5  &85.5 \\
    BoP \cite{Lin2012}     &87.8 &99.4  &99.8 \\
    Proposed method &\bm{$93.8$}  &\bm{$99.5$}   &\bm{$100$} \\
  \bottomrule
  \end{tabular}
\end{table}

\begin{table*} [!t]
 \renewcommand{\arraystretch}{1.3}
  \caption{Comparison of the classification accuracy (\%) on the epileptic EEG dataset.}
  \label {comparision}
  \centering
  \begin {tabular} {l l c l r}
  \toprule[1pt]
   \bfseries Researchers &\bfseries Datasets &\bfseries Num of Class &\bfseries Methods &\bfseries Accuracy (\%)\\ \midrule

    Kannathal et al. \cite{Kannathal2005}         &A, E             &2 &Entropy+adaptive neuro-fuzzy inference system    &95\\ \midrule

    Polat and G\"unes \cite{Polat2007}         &A, E             &2 &FFT+decision tree    &98.72\\ \midrule

    Yuan et al. \cite{Yuan2011}         &D, E             &2 &Nonlinear features+extreme learning machine  &96.00\\ \midrule

    Ocak \cite{Hasan2009}         &(A, C, D), E             &2 &DWT+approximate entropy  &96.65\\ \midrule

    Guo et al. \cite{Guo2010}         &A, E             &2 &Line length feature based on DWT+artificial neural networks    &99.6\\
                        &(A, C, D), E      &2 &Line length feature based on DWT+artificial neural networks    &97.75\\
                        &(A, B, C, D), E   &2 &Line length feature based on DWT+artificial neural networks    &97.77\\ \midrule

    G\"uler and \"Ubeyli \cite{I.Guler2007}        &A, B, C, D, E      &5 &Raw Data+support vector machine    &75.6\\
              &       &  &Raw Data+probabilistic neural network      &72.00\\
              &       &  &Raw Data+MLPNN    &68.80\\
              &       &  &DWT and lyapunov exponents+support vector machine    &99.28\\
              &       &  &DWT and lyapunov exponents+probabilistic neural network    &98.05\\
              &       &  &DWT and lyapunov exponents+MLPNN  &93.63\\ \midrule

    This work             &    A, E        &2   &Bag-of-words+ 1-NN & 99.5\\
                     &    (A, C, D), E        &2   &Bag-of-words+ 1-NN & 99.0\\
                     &    (A, B, C, D), E        &2   &Bag-of-words+ 1-NN & 99.2\\
                  &    A, B, C, D, E        &5   &Bag-of-words+ 1-NN & 93.8\\
  \bottomrule
  \end{tabular}
\end{table*}

We compared the performance of the proposed bag-of-words representation with that of the DWT representation \cite{.I.Guler2005}, the DFT representation \cite{Rafiei2000}, and the NN classifier based on the DTW distance \cite{Fu2011}. In addition, we also compared the proposed bag-of-words representation with the bag-of-patterns representation \cite{Lin2012}, which is very similar to the proposed approach.

\begin{itemize}
\setlength{\itemsep}{1pt}
\setlength{\parskip}{0pt}
\setlength{\parsep}{0pt}
\item DWT that represents a signal in multiresolution is able to capture both frequency and location information of time series. Similar to the DWT based feature used in \cite{.I.Guler2005}, we used the Daubechies wavelet (db2) and decomposed the time series into 4 levels. The detail wavelet coefficients of the four levels and the approximation wavelet coefficients of the fourth level are concatenated to form the final representation.

\item DFT is a widely used transformation technique to extract frequency information from time series. We transformed the original time series into the frequency domain and extracted the DFT coefficients as features.

\item DTW that uses dynamic programming technique to determine the best alignment of two sequences is able to deal with temporal drift between time series. The distance matrix of each pair of the test time series and the training time series is calculated based on the unconstrained DTW. This distance matrix is used as input of the NN classifier.

\item The BoP representation that represents a time series as a histogram of local patterns is very similar to the proposed bag-of-words representation. The size of alphabet $\tau$ and the number of symbols $w$ are empirically set to 4 and 6, respectively. We varied the length of local segments in the bag-of-patterns representation from 16 to 320 with a step of 16. The best accuracy is reported for comparison.

\end{itemize}

Since the time series in the ECG-15 datasets have different lengths (2048$\sim$4096), we resized all the time series to the same length of 4096 using bilinear interpolation so that the DWT and DFT based features have the same dimension. When calculating the DTW distance, we reduced all the time series in the three datasets to the length of 820 with a downsampling rate of about 5 because DTW is computationally expensive.

Table \ref{methods} summarizes the best results achieved by the proposed approach and the other methods. It can be seen that the proposed approach achieves the highest accuracies (93.8\% on the EEG dataset, 99.5\% on the ECG-40 dataset, and 100\% on the ECG-15 dataset, respectively), which illustrate the effectiveness of the bag-of-words representation. The BoP representation obtains comparable accuracies on the ECG-40 and the ECG-15 datasets with that by the bag-of-words representation. However, the proposed bag-of-words representation performs significantly better than the BoP representation on the EEG dataset. The DFT feature and DTW distance methods outperform the DWT based method. This is probably because that the DFT and DTW can better deal with temporal sift between sequences than the DWT.

The EEG dataset used in our experiment is a popular dataset for automatic epileptic seizure classification and localization. Table \ref{comparision} provides a comparison of the classification accuracies between the proposed bag-of-words method and previous state-of-the-art approaches in the literature. It should be noticed that the comparison is not direct, since the aim of our method is to classify the time series at sequence level, while the other methods are to classify segments extracted from the time series. Some works used only several subsets of the whole EEG dataset to construct a 2-class dataset, while others used the whole EEG dataset with 5 classes. For the 2-class classification, the bag-of-words method outperforms most of the other methods. For the 5-class classification where the whole EEG dataset is used, the classification accuracies of support vector machine (SVM), probabilistic neural network (PNN) and multilayer perception neural network (MLPNN) with raw data are 75.60\% 72.00\% and 68.80\% \cite{I.Guler2007}, respectively. When features extracted from DWT and lyapunov exponents are used, the corresponding accuracies increase to 99.28\%, 98.05\% and 93.63\% \cite{I.Guler2007}, respectively. The result obtained by the proposed BoW representation with the simplest NN classifier is slightly lower than those achieved by SVM and PNN with features based on DWT and lyapunov exponents. However, it is slightly higher than the result obtained by MLPNN (93.63\%) with features based on DWT and lyapunov exponents.

\subsection{Robustness to Noise}
\begin{table} [!t]
 \renewcommand{\arraystretch}{1.3}
  \caption{Classification accuracies (\%) on the three datasets corrupted by zero-mean white Gaussian noise.}
  \label {Noise}
  \centering
  \begin {tabular} {l c c c}
  \toprule[1pt]
   \bfseries SNR &\bfseries EEG &\bfseries ECG-40 &\bfseries ECG-15 \\ \midrule
    10db &92.6  &98.9   &99.8 \\
    8db  &91.8  &98.4   &99.7 \\
    6db  &91.2  &97.6   &99.6 \\
    4db  &90.4  &95.5   &99.2 \\
    2db  &88.8  &92.6   &98.9 \\
    0db  &85.2  &89.9   &98.6 \\
  \bottomrule
  \end{tabular}
\end{table}
This experiment is designed to investigate the robustness of the bag-of-words representation to noise. All signals in the EEG, ECG-40 and ECG-15 datasets were corrupted by zero mean white Gaussian noise. The standard deviation of the white Gaussian noise is varied so that the SNRs are between 10dB and 0dB. The training data and the test data are separated exactly the same as those in the previous experiments. Table \ref{Noise} summaries the classification accuracies on the three datasets contaminated by the white Gaussian noise with different SNRs. It can be seen that the bag-of-words approach is relatively robust to noise. The accuracies decreased by less than 2 percents when the SNR is 10dB. Even for considerable noise contamination with the SNR 0dB, the accuracies reduced less than 10 percents for the EEG and ECG-40 datasets, and only less than 2 percents for the ECG-15 dataset.

\section{Discussion}
\label{sec:discussion}
Although the bag-of-words representation ignores the temporal order of local segments, it is able to effectively capture high-level structural information due to the fact that the frequency of the codewords (local segments) occurred in a time series is well utilized. However, since the local segments are extracted by sliding a window along time series, a time series that is not reasonably long cannot provide enough local segments to capture local structures in the time series. Therefore, the bag-of-words representation may be ineffective to represent short time series, which is mainly due to the limitation that the bag-of-words representation cannot extract enough meaningful and discriminative local segments from short sequences.

The size of the codebook $N$ is pre-defined and empirically determined in the method. A compact codebook with small size has a limited discriminative ability, while a codebook with large size is likely to introduce noise. How to adaptively set the optimal size of the codebook to make the codebook compact and yet discriminative is still an open question. Some criteria can be defined to merge entries of a codebook to construct an adaptive codebook. For instance, the method in \cite{Liu2008} utilized Maximization of Mutual Information (MMI) principal to estimate the optimal $N$. Two entries of a codebook are merged by maximizing the mutual information in an unsupervised way. Creating a codebook with adaptive size will be investigated in our future work.

\section{Conclusion}
\label{sec:conclusion}
In this paper, we proposed a bag-of-words representation for biomedical time series analysis. The proposed method treats a time series as a document and local segments extracted from the time series as words. The time series is represented as a histogram of codewords. Although the temporal order information of the local segments is ignored, both local structure and global structure information of the time series are captured. Experimental results on three publicly available datasets demonstrate that the bag-of-words representation is effective for characterizing biomedical time series such as EEG and ECG signals. Furthermore, the bag-of-words representation is not only insensitive to the length of local segments and the size of codebook, but also robust to noise. The distance measures for comparison of histograms are also investigated in the experiments, showing that the Chi-Squared distance measure is more suitable for comparing histograms than the other distance measures. We compared the performance of the bag-of-words representation with several state-of-the-art approaches in the literature. Experimental results show that the bag-of-words representation with the simplest 1-Nearest Neighbor (1-NN) classifier achieves comparable or higher classification accuracies than those by the others.


%

%

%

\ifCLASSOPTIONcaptionsoff
  \newpage
\fi



%


\bibliographystyle{IEEEtran}
\bibliography{BioTimeSeries}

\begin{thebibliography}{10}
\providecommand{\url}[1]{#1}
\csname url@samestyle\endcsname
\providecommand{\newblock}{\relax}
\providecommand{\bibinfo}[2]{#2}
\providecommand{\BIBentrySTDinterwordspacing}{\spaceskip=0pt\relax}
\providecommand{\BIBentryALTinterwordstretchfactor}{4}
\providecommand{\BIBentryALTinterwordspacing}{\spaceskip=\fontdimen2\font plus
\BIBentryALTinterwordstretchfactor\fontdimen3\font minus
  \fontdimen4\font\relax}
\providecommand{\BIBforeignlanguage}[2]{{%
\expandafter\ifx\csname l@#1\endcsname\relax
\typeout{** WARNING: IEEEtran.bst: No hyphenation pattern has been}%
\typeout{** loaded for the language `#1'. Using the pattern for}%
\typeout{** the default language instead.}%
\else
\language=\csname l@#1\endcsname
\fi
#2}}
\providecommand{\BIBdecl}{\relax}
\BIBdecl

\bibitem{Kannathal2005}
N.~Kannathal, M.~L. Choo, U.~R. Acharya, and P.~Sadasivan, ``Entropies for
  detection of epilepsy in {EEG},'' \emph{Compu. Meth. Prog. Bio.}, vol.~80,
  no.~3, pp. 187 -- 194, 2005.

\bibitem{Guo2010}
L.~Guo, D.~Rivero, J.~Dorado, J.~R. Rabu–al, and A.~Pazos, ``Automatic
  epileptic seizure detection in {EEG}s based on line length feature and
  artificial neural networks,'' \emph{J. Neurosci Meth.}, vol. 191, no.~1, pp.
  101 -- 109, 2010.

\bibitem{Yuan2011}
Q.~Yuan, W.~Zhou, S.~Li, and D.~Cai, ``Epileptic {EEG} classification based on
  extreme learning machine and nonlinear features,'' \emph{Epilepsy Res.},
  vol.~96, no. 1-2, pp. 29 -- 38, 2011.

\bibitem{Wolpaw2002}
J.~R. Wolpaw, N.~Birbaumer, D.~J. McFarland, G.~Pfurtscheller, and T.~M.
  Vaughan, ``Brain computer interfaces for communication and control,''
  \emph{Clin. Neurophysiol.}, vol. 113, no.~6, pp. 767 -- 791, 2002.

\bibitem{Lu2010}
H.~Lu, H.-L. Eng, C.~Guan, K.~Plataniotis, and A.~Venetsanopoulos,
  ``Regularized common spatial pattern with aggregation for {EEG}
  classification in small-sample setting,'' \emph{IEEE Trans. Biomed. Eng.},
  vol.~57, no.~12, pp. 2936 --2946, dec. 2010.

\bibitem{Wang2011}
H.~Wang, ``Multiclass filters by a weighted pairwise criterion for {EEG}
  single-trial classification,'' \emph{IEEE Trans. Biomed. Eng.}, vol.~58,
  no.~5, pp. 1412 --1420, may 2011.

\bibitem{Shen2007}
K.-Q. Shen, C.-J. Ong, X.-P. Li, Z.~Hui, and E.~Wilder-Smith, ``A feature
  selection method for multilevel mental fatigue {EEG} classification,''
  \emph{IEEE Trans. Biomed. Eng.}, vol.~54, no.~7, pp. 1231 --1237, july 2007.

\bibitem{Lin2010}
Y.-P. Lin, C.-H. Wang, T.-P. Jung, T.-L. Wu, S.-K. Jeng, J.-R. Duann, and J.-H.
  Chen, ``{EEG}-based emotion recognition in music listening,'' \emph{IEEE
  Trans. Biomed. Eng.}, vol.~57, no.~7, pp. 1798 --1806, july 2010.

\bibitem{Ince2009}
T.~Ince, S.~Kiranyaz, and M.~Gabbouj, ``A generic and robust system for
  automated patient-specific classification of {ECG} signals,'' \emph{IEEE
  Trans. Biomed. Eng.}, vol.~56, no.~5, pp. 1415 --1426, may 2009.

\bibitem{Kampouraki2009}
A.~Kampouraki, G.~Manis, and C.~Nikou, ``Heartbeat time series classification
  with support vector machines,'' \emph{IEEE Trans. Inf. Technol. Biomed.},
  vol.~13, no.~4, pp. 512 --518, july 2009.

\bibitem{Huken2003}
M.~Huken and P.~Stagge, ``Recurrent neural networks for time series
  classification,'' \emph{Neurocomputing}, vol.~50, pp. 223 -- 235, 2003.

\bibitem{Zadeh2010}
A.~E. Zadeh, A.~Khazaee, and V.~Ranaee, ``Classification of the
  electrocardiogram signals using supervised classifiers and efficient
  features,'' \emph{Compu. Meth. Prog. Bio.}, vol.~99, no.~2, pp. 179 -- 194,
  2010.

\bibitem{.I.Guler2005}
\.I.G\"uler and E.~D. \"Ubeyli, ``{ECG} beat classifier designed by combined
  neural network model,'' \emph{Pattern Recogn.}, vol.~38, no.~2, pp. 199 --
  208, 2005.

\bibitem{Lin2012}
J.~Lin, R.~Khade, and Y.~Li, ``Rotation-invariant similarity in time series
  using bag-of-patterns representation,'' \emph{J. Intell. Inf. Syst}, pp.
  1--29, 2012.

\bibitem{Lebanon2007}
G.~Lebanon, Y.~Mao, and J.~Dillon, ``The locally weighted bag of words
  framework for document representation,'' \emph{J. Mach. Learn. Res.}, vol.~8,
  pp. 2405--2441, December 2007.

\bibitem{Blei2003}
D.~Blei, A.~Ng, and M.~Jordan, ``Latent dirichlet allocation,'' \emph{J. Mach.
  Learn. Res.}, vol.~3, pp. 993--1022, 2003.

\bibitem{Fei-Fei2005}
L.~Fei-Fei and P.~Perona, ``A bayesian hierarchical model for learning natural
  scene categories,'' in \emph{Proc. IEEE Int'l Conf. Computer Vision and
  Pattern Recognition}, vol.~2, 2005, pp. 524 -- 531.

\bibitem{Niebles2008}
J.~C. Niebles, H.~Wang, and L.~Fei-Fei, ``Unsupervised learning of human action
  categories using spatial-temporal words,'' \emph{Int. J. Comput. Vision.},
  vol.~79, no.~3, pp. 299--318, 2008.

\bibitem{Jurie2005}
F.~Jurie and B.~Triggs, ``Creating efficient codebooks for visual
  recognition,'' in \emph{Proc. IEEE Int'l Conf. Computer Vision}, vol.~1, oct.
  2005, pp. 604 -- 610 Vol. 1.

\bibitem{Fernando2012}
B.~Fernando, E.~Fromont, D.~Muselet, and M.~Sebban, ``Supervised learning of
  gaussian mixture models for visual vocabulary generation,'' \emph{Pattern
  Recogn.}, vol.~45, no.~2, pp. 897 -- 907, 2012.

\bibitem{I.Guler2007}
\.I.G\"uler and E.~D. \"Ubeyli, ``Multiclass support vector machines for
  {EEG}-signals classification,'' \emph{IEEE Trans. Inf. Technol. Biomed.},
  vol.~11, no.~2, pp. 117 --126, march 2007.

\bibitem{Endres2003}
D.~Endres and J.~Schindelin, ``A new metric for probability distributions,''
  \emph{IEEE Trans. Inf. Theory}, vol.~49, no.~7, pp. 1858 -- 1860, july 2003.

\bibitem{Grauman2007}
K.~Grauman and T.~Darrell, ``The pyramid match kernel: Efficient learning with
  sets of features,'' \emph{J. Mach. Learn. Res.}, vol.~8, pp. 725--760, May
  2007.

\bibitem{Biel2001}
L.~Biel, O.~Pettersson, L.~Philipson, and P.~Wide, ``{ECG} analysis: a new
  approach in human identification,'' \emph{IEEE Trans. Instrum. Meas.},
  vol.~50, no.~3, pp. 808 --812, 2001.

\bibitem{Fang2009}
S.-C. Fang and H.-L. Chan, ``Human identification by quantifying similarity and
  dissimilarity in electrocardiogram phase space,'' \emph{Pattern Recogn.},
  vol.~42, no.~9, pp. 1824 -- 1831, 2009.

\bibitem{Andrzejak2001}
R.~G. Andrzejak, K.~Lehnertz, F.~Mormann, C.~Rieke, P.~David, and C.~E. Elger,
  ``Indications of nonlinear deterministic and finite-dimensional structures in
  time series of brain electrical activity: dependence on recording region and
  brain state.'' \emph{Phys. rev. E}, vol.~64, no. 6 Pt 1, Dec. 2001.

\bibitem{Goldberger2000}
A.~L. Goldberger, L.~A.~N. Amaral, L.~Glass, J.~M. Hausdorff, P.~Ch.Ivanov,
  R.~G. Mark, J.~E. Mietus, G.~B. Moody, C.-K. Peng, and H.~E. Stanley,
  ``Physio{B}ank, {P}hysio{T}oolkit, and {P}hysio{N}et: components of a new
  research resource for complex physiologic signals,'' \emph{Circulation}, vol.
  101, no.~23, pp. 215--220, 2000.

\bibitem{Rafiei2000}
D.~Rafiei and A.~Mendelzon, ``Querying time series data based on similarity,''
  \emph{IEEE Trans. Knowl. Data Eng.}, vol.~12, no.~5, pp. 675 --693, 2000.

\bibitem{Fu2011}
T.~chung Fu, ``A review on time series data mining,'' \emph{Eng. Appl. Artif.
  Intel.}, vol.~24, no.~1, pp. 164 -- 181, 2011.

\bibitem{Polat2007}
K.~Polat and S.~G\"unes, ``Classification of epileptiform {EEG} using a hybrid
  system based on decision tree classifier and fast fourier transform,''
  \emph{Appl. Math. Comput.}, vol. 187, no.~2, pp. 1017 -- 1026, 2007.

\bibitem{Hasan2009}
H.~Ocak, ``Automatic detection of epileptic seizures in {EEG} using discrete
  wavelet transform and approximate entropy,'' \emph{Expert Syst. Appl.},
  vol.~36, no. 2, Part 1, pp. 2027 -- 2036, 2009.

\bibitem{Liu2008}
J.~Liu and M.~Shah, ``Learning human actions via information maximization,'' in
  \emph{Proc. IEEE Int'l Conf. Computer Vision and Pattern Recognition}, 2008,
  pp. 1--8.

\end{thebibliography}

%

\begin{biography}[{\includegraphics[width=1in,height=1.25in,clip,keepaspectratio]{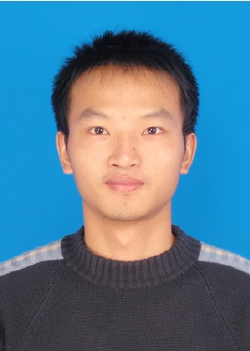}}]{Jin Wang}
received his M.S. in Pattern Recognition and Artificial Intelligence from Huazhong University of Science and Technology, China. Currently, he is pursuing the Ph.D. degree in the Center for Intelligent System Research (CISR) and the Institute for Frontier Material (IFM) at Deakin University, Australia. His major research interests are signal processing, machine learning and intelligent system, including biomedical time series analysis, automatic image and video analysis, and intelligent wearable systems.
\end{biography}

\begin{biography}[{\includegraphics[width=1in,height=1.25in,clip,keepaspectratio]{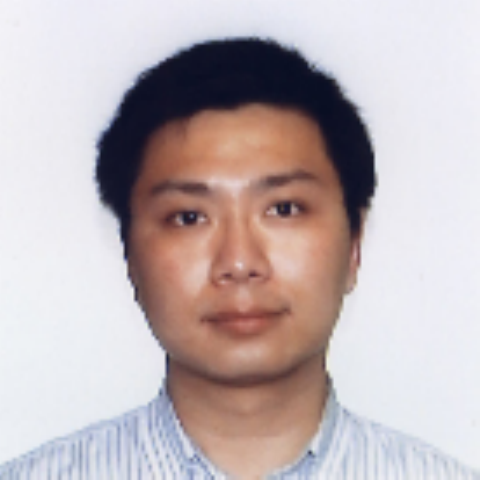}}]{Ping Liu}
got his Bachelor degree (EE) in WuHan university of Technology, 2005; Master degree (CS) in Huazhong University of Science and Technology, 2008. From 2011, he have been studying in the University of South Carolina, American. Generally, his interest includes computer vision, pattern recognition and machine learning.
\end{biography}

\begin{biography}[{\includegraphics[width=1in,height=1.25in,clip,keepaspectratio]{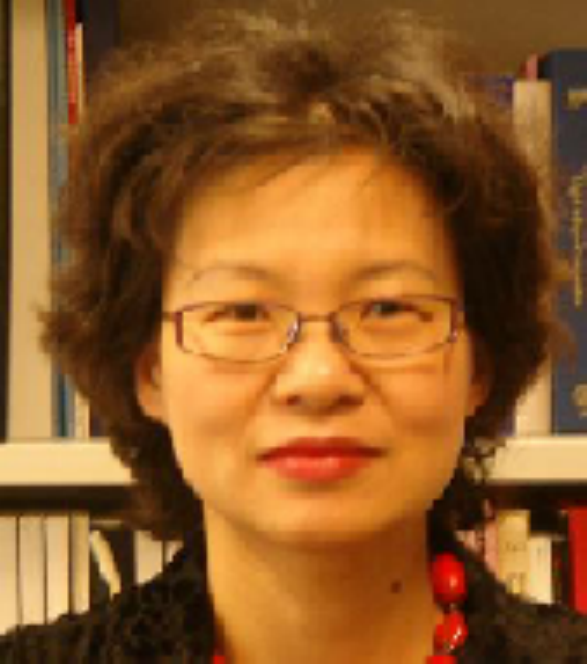}}]{Mary F.H.She}
received her B.Sc. and M.Sc. degrees in Engineering from Donghua University, Shanghai, China and Ph.D from Deakin University, Victoria, Australia.  She was awarded Australian Post-doctorial Fellowship by Australian Research Council in 2002.  She currently is a research fellow in the Center for Intelligent System Research (CISR) and the Institute for Frontier Material (IFM) at Deakin University.  Her major research interest includes image processing and intelligent wearable systems.
\end{biography}

\begin{biography}[{\includegraphics[width=1in,height=1.25in,clip,keepaspectratio]{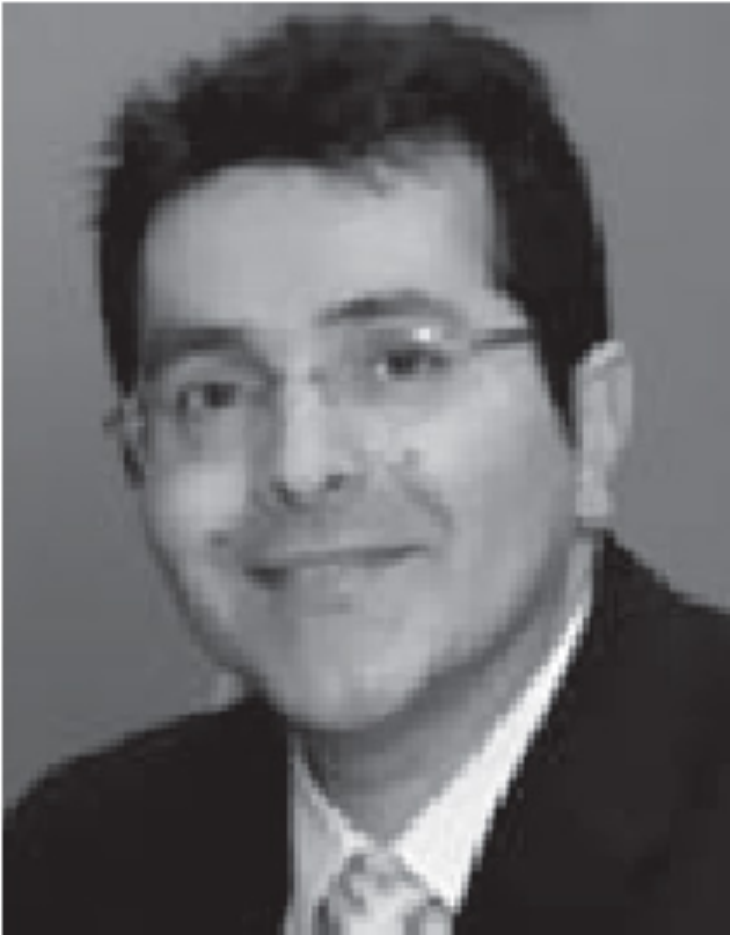}}]{Saeid Nahavandi}
received the B.Sc. (hons.), M.Sc., and Ph.D. degrees in automation and control from Durham University, Durham, U.K. He is the Alfred Deakin Professor, Chair of Engineering, and the director for the Center for Intelligent Systems Research (CISR), Deakin University, Geelong, VIC, Australia. His research interests include modeling of complex systems, simulation-based optimization, robotics, haptics and augmented reality. Dr. Nahavandi is the Associate Editor of the IEEE Systems Journal, an Editorial Consultant Board member for the International Journal of Advanced Robotic Systems, an Editor (South Pacific Region) of the International Journal of Intelligent Automation and Soft Computing. He is a Fellow of Engineers Australia (FIEAust), IET (FIET) and Senior member of IEEE (SMIEEE).
\end{biography}

\begin{biography}[{\includegraphics[width=1in,height=1.25in,clip,keepaspectratio]{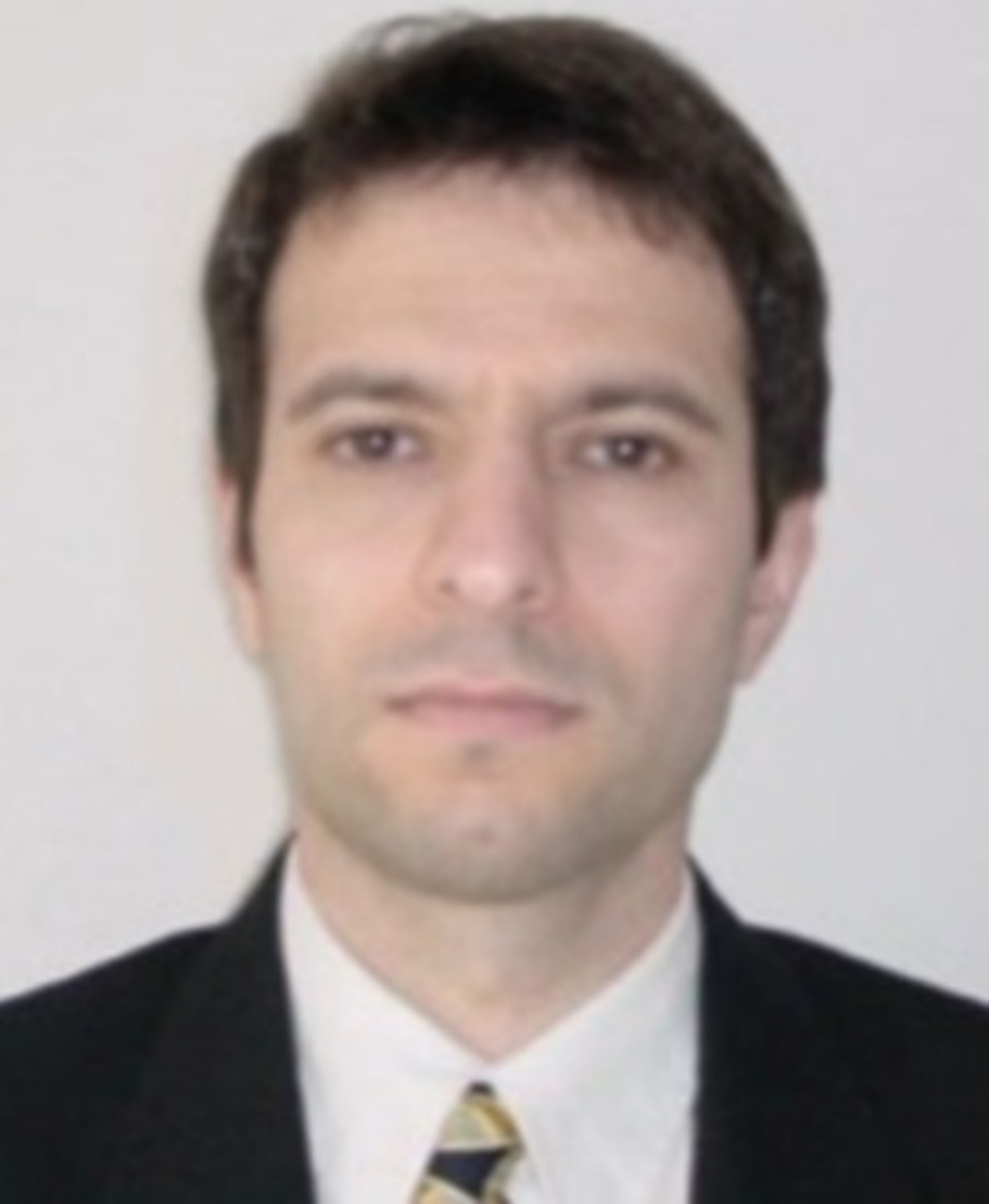}}]{Abbas Kouzani}
received his M.Eng. degree in electrical and electronics engineering from the University of Adelaide, Australia, 1995, and Ph.D. degree in electrical and electronics engineering from Flinders University, Australia, 1999. Currently, he is an Associate Professor with the School of Engineering, Deakin University. He has been involved in several ARC and industry research grants, and more than 150 publications. His research interests include pattern recognition and intelligent systems.
\end{biography}




\end{document}